\documentclass[a4paper]{article}

\usepackage[english]{babel}
\usepackage[utf8]{inputenc}

\usepackage[a4paper, total={6.5in, 9in}]{geometry}
\usepackage[backend=bibtex,style=numeric-comp,giveninits=true,doi=false,isbn=false,url=false,eprint=false,maxbibnames=99]{biblatex}

\usepackage{amsmath}
\usepackage{amssymb}
\usepackage{mathtools}
\usepackage{mathrsfs}

\usepackage{siunitx}
\usepackage[colorinlistoftodos]{todonotes}
\usepackage[toc,page]{appendix}
\usepackage[hidelinks]{hyperref}

\usepackage{tabularx}
\usepackage{booktabs}
\usepackage{multirow}

\usepackage{rotating}

\usepackage{amsthm}
\theoremstyle{plain}
\theoremstyle{plain}
\theoremstyle{plain}
\theoremstyle{plain}
\theoremstyle{plain}
\theoremstyle{definition}
\theoremstyle{remark}\newtheorem{remark}{Remark}
\newcommand{\papertitle}{Latent Dynamics Networks (LDNets): learning the intrinsic dynamics of spatio-temporal processes}

\newcommand{\figpt}[1]{\textbf{(#1)}}

\newcommand{\avsum}{\mathop{\mathpalette\avsuminner\relax}\displaylimits}

\makeatletter
\newcommand\avsuminner[2]{%
  {\sbox0{$\m@th#1\sum$}%
   \vphantom{\usebox0}%
   \ooalign{%
     \hidewidth
     \smash{\vrule height\dimexpr\ht0+1pt\relax depth\dimexpr\dp0+1pt\relax}%
     \hidewidth\cr
     $\m@th#1\sum$\cr
   }%
  }%
}
\makeatother

\newcommand{\paramsgeneric}{\mathbf{w}}

\newcommand{\stateROM}{\mathbf{s}}
\newcommand{\TARGETstateROM}{\hat{\mathbf{s}}}
\newcommand{\stateFOM}{\mathbf{z}}

\newcommand{\inputSign}{\mathbf{u}}

\newcommand{\outputField}{\mathbf{y}}

\newcommand{\DISCRFOMrhs}{\mathbf{F}}
\newcommand{\statePROM}{\stateROM}
\newcommand{\statePROMcoeff}{s_{k}}
\newcommand{\statePROMcoeffIdx}[1]{s_{#1}}

\newcommand{\NUMsnapshots}{N_{\mathrm{snapshots}}}

\newcommand{\EQinputSign}{\inputSign_{\mathrm{eq}}}

\newcommand{\DISCRstateFOM}{\mathbf{Z}}
\newcommand{\DISCRoutputField}{\mathbf{Y}}
\newcommand{\APPDISCRoutputField}{\widetilde{\mathbf{Y}}}

\newcommand{\CENTERinputSign}{\mathbf{u}_0}
\newcommand{\CENTERoutputField}{\mathbf{y}_0}
\newcommand{\CENTERspacevar}{\mathbf{x}_0}
\newcommand{\WIDTHinputSign}{\mathbf{u}_w}
\newcommand{\WIDTHoutputField}{\mathbf{y}_w}
\newcommand{\WIDTHspacevar}{\mathbf{x}_w}

\newcommand{\NORMoutputField}{y_{\mathrm{norm}}}

\newcommand{\spacevar}{\mathbf{x}}
\newcommand{\timevar}{t}
\newcommand{\OBSspacevar}{\boldsymbol{\xi}}
\newcommand{\OBStimevar}{\tau}

\newcommand{\Dt}{\Delta t}
\newcommand{\DtREF}{\Delta t_{\mathrm{ref}}}

\newcommand{\APPoutputField}{\widetilde{\outputField}}

\newcommand{\spaceDomain}{\Omega}
\newcommand{\timeMax}{T}

\newcommand{\SPACEstateFOM}{Z}

\newcommand{\SPACEinputSign}{U}
\newcommand{\SPACEoutputField}{Y}

\newcommand{\SETstateFOM}{\mathcal{Z}}

\newcommand{\SETinputSign}{\mathcal{U}}

\newcommand{\SEToutputField}{\mathcal{Y}}

\newcommand{\NUMstateROM}{d_{\stateROM}}
\newcommand{\NUMstateFOM}{d_{\stateFOM}}
\newcommand{\NUMstateDISCR}{N_h}
\newcommand{\NUMobsPTS}{N_{\mathrm{nodes}}}

\newcommand{\NUMinputSign}{d_{\inputSign}}

\newcommand{\NUMoutputField}{d_{\outputField}}
\newcommand{\NUMspacevar}{d}

\newcommand{\trainTIMESinputSign}[1]  {\mathscr{S}^{#1}}
\newcommand{\trainTIMESoutputField}[1]{\mathscr{T}^{#1}}
\newcommand{\trainPOINTSoutputField}[2]{\mathscr{P}^{#1}_{#2}}

\newcommand{\NNgeneric}{\mathcal{N\!N}}
\newcommand{\RNNgeneric}{\mathcal{R\!N\!N}}
\newcommand{\FOMrhs}{\mathcal{F}}
\newcommand{\FOMobs}{\mathcal{G}}

\newcommand{\ROMrhs}{\NNgeneric_{\mathrm{dyn}}}
\newcommand{\ROMobs}{\NNgeneric_{\mathrm{rec}}}
\newcommand{\ROMdyn}{\RNNgeneric_{\mathrm{dyn}}}
\newcommand{\ROMenc}{\NNgeneric_{\mathrm{enc}}}
\newcommand{\ROMdec}{\NNgeneric_{\mathrm{dec}}}

\newcommand{\adimROMrhs}{\widetilde{\NNgeneric}_{\mathrm{dyn}}}
\newcommand{\adimROMobs}{\widetilde{\NNgeneric}_{\mathrm{rec}}}

\newcommand{\paramsROMdyn}{\paramsgeneric_{\mathrm{dyn}}}
\newcommand{\paramsROMobs}{\paramsgeneric_{\mathrm{rec}}}
\newcommand{\paramsROMenc}{\paramsgeneric_{\mathrm{enc}}}
\newcommand{\paramsROMdec}{\paramsgeneric_{\mathrm{dec}}}

\newcommand{\trainSamples}{\mathcal{S}_{\mathrm{train}}}
\newcommand{\testSamples}{\mathcal{S}_{\mathrm{test}}}

\newcommand{\discrOperatorCD}{\mathcal{A}}
\newcommand{\discrOperatorDC}{\mathcal{A}'}

\newcommand{\loss}{\mathcal{L}}
\newcommand{\error}{\mathcal{E}}
\newcommand{\lossREG}{\mathcal{R}}
\newcommand{\weightREGrhs}{\alpha_{\mathrm{dyn}}}
\newcommand{\weightREGobs}{\alpha_{\mathrm{rec}}}
\newcommand{\weightREGenc}{\alpha_{\mathrm{enc}}}
\newcommand{\weightREGdec}{\alpha_{\mathrm{dec}}}

\newcommand{\variable}{\alpha}
\newcommand{\ADIMvariable}{\tilde{\alpha}}
\newcommand{\CENTERvariable}{\variable_0}
\newcommand{\WIDTHvariable}{\variable_w}
\newcommand{\MINvariable}{\variable_{\min}}
\newcommand{\MAXvariable}{\variable_{\max}}

\newcommand{\VECinputSign}{\{ \inputSign_i(\OBStimevar) \}_{\OBStimevar \in \trainTIMESinputSign{i}}}

\newcommand{\DirichletDatum}{\outputField_{D}}
\newcommand{\DirichletLifting}{\outputField_{\mathrm{lift}}}
\newcommand{\DirichletMask}{\psi}

\newcommand{\ADRstate}{z}
\newcommand{\fmin}{f_{\mathrm{min}}}
\newcommand{\fmax}{f_{\mathrm{max}}}

\newcommand{\velocity}{\mathbf{v}}
\newcommand{\GammaTop}{\Gamma_{\mathrm{top}}}
\newcommand{\NSrefVel}{v_{\mathrm{norm}}}

\newcommand{\APstate}{z}
\newcommand{\APrecovery}{w}
\newcommand{\APL}{L}
\newcommand{\APD}{D}
\newcommand{\APK}{K}
\newcommand{\APalpha}{\alpha}
\newcommand{\APgamma}{\gamma}
\newcommand{\APmuOne}{\mu_1}
\newcommand{\APmuTwo}{\mu_2}
\newcommand{\APb}{b}
\newcommand{\APIapp}{I_{\mathrm{stim}}}
\newcommand{\APxapp}[1]{x^{\mathrm{stim}}_{#1}}

\graphicspath{{img/}}

\title{{\papertitle}}
\author{Francesco Regazzoni$^{1,*}$,
		Stefano Pagani$^1$,
		Matteo Salvador$^{1, 2}$,
		Luca Dede'$^1$,
		Alfio Quarteroni$^{1, 3}$}

\date{\footnotesize
	$^1$ MOX, Department of Mathematics, Politecnico di Milano, Milan, Italy \\
	$^2$ Institute for Computational and Mathematical Engineering, Stanford University, California, USA \\
	$^3$ \'Ecole Polytechnique F\'ed\'erale de Lausanne, Lausanne, Switzerland (\textit{Professor Emeritus})\\[2ex]
	$^*$ \textit{Corresponding author} (\texttt{francesco.regazzoni@polimi.it}) \\
    }

\begin{document}
	\maketitle

	\begin{abstract}
		Predicting the evolution of systems that exhibit spatio-temporal dynamics in response to external stimuli is a key enabling technology fostering scientific innovation. Traditional equations-based approaches leverage first principles to yield predictions through the numerical approximation of high-dimensional systems of differential equations, thus calling for large-scale parallel computing platforms and requiring large computational costs. Data-driven approaches, instead, enable the description of systems evolution in low-dimensional latent spaces, by leveraging dimensionality reduction and deep learning algorithms. We propose a novel architecture, named Latent Dynamics Network (LDNet), which is able to discover low-dimensional intrinsic dynamics of possibly non-Markovian dynamical systems, thus predicting the time evolution of space-dependent fields in response to external inputs. Unlike popular approaches, in which the latent representation of the solution manifold is learned by means of auto-encoders that map a high-dimensional discretization of the system state into itself, LDNets automatically discover a low-dimensional manifold while learning the latent dynamics, without ever operating in the high-dimensional space. Furthermore, LDNets are meshless algorithms that do not reconstruct the output on a predetermined grid of points, but rather at any point of the domain, thus enabling weight-sharing across query-points. These features make LDNets lightweight and easy-to-train, with excellent accuracy and generalization properties, even in time-extrapolation regimes. We validate our method on several test cases and we show that, for a challenging highly-nonlinear problem, LDNets outperform state-of-the-art methods in terms of accuracy (normalized error 5 times smaller), by employing a dramatically smaller number of trainable parameters (more than 10 times fewer).
	\end{abstract}

	\section{Introduction}\label{sec:PP:introduction}

Mathematical models based on differential equations, such as Partial Differential Equations (PDEs) and Stochastic Differential Equations (SDEs), can yield quantitative predictions of the evolution of space-dependent quantities of interest in response to external stimuli. 
Pivotal examples are given by fluid dynamics and turbulence \cite{patera1984spectral}, wave propagation phenomena \cite{monaghan1994simulating}, the deformation of solid bodies and biological tissues \cite{regazzoni2022emcirculation}, molecular dynamics \cite{bird1994molecular}, price evolution of financial assets \cite{scalas2000fractional}, epidemiology \cite{lai2020effect}.
However, the development of traditional modeling-and-simulation approaches carry several mathematical and computational challenges.
Model development requires a deep understanding of the physical processes, the adoption of physics first principles or empirical rules, and their translation into mathematical terms. 
The values of parameters and of boundary and initial conditions required to close the model are often unknown, increasing the intrinsic dimensionality of the solution space. 
Finally, the computational cost that accompanies the (possibly many-query) numerical approximation of such mathematical models may be prohibitive and hinder their use in relevant applications \cite{quarteroni2015reduced,peherstorfer2015dynamic}.

In recent years, we are witnessing the introduction of a new paradigm, namely data-driven modeling \cite{bongard2007automated,schmidt2009distilling,peherstorfer2017data,rudy2017data,bar2019learning,cenedese2022data}, as opposed to traditional physics-based modeling, enabled by recent advances in optimization, high-performance computing, GPU-based hardware, artificial neural networks (NNs) and Machine/Deep Learning in general.
Data-driven modeling methods hold promise in overcoming the limitations of traditional physics-based models, either as a replacement for them or in synergy with them \cite{alber2019integrating}. 
On the one hand, data-driven techniques are employed to learn a model directly from experimental data \cite{bongard2007automated,schmidt2009distilling}. 
On the other hand, instead, they are used to build a surrogate for a high-fidelity model -- the latter being typically based on the numerical approximation of systems of differential equations -- from a dataset of precomputed high-fidelity simulation snapshots \cite{bar2019learning,alber2019integrating}. 
This paradigm is successful in many-query contexts, that is when the computational resources spent in the offline phase (generation of the training data and construction of the data-driven surrogate model) are repaid by a large number of evaluations of the trained model (online phase), as is the case of sensitivity analysis, parameter estimation and uncertainty quantification. 
Another case of interest is when real-time responses are needed, like, e.g., in clinical scenarios \cite{peirlinck2021precision}.

Several methods have been recently proposed for automatically learning the dynamics of systems exhibiting spatio-temporal behavior \cite{rudy2017data,oommen2022learning,vlachas2022multiscale,floryan2022data}.
Typically, these methods discretize the space-dependent output field into a high-dimensional vector (e.g., by point-wise evaluation on a grid, or by expansion with respect to a Finite Element basis or to a Fourier basis) and then compress it by means of dimensionality reduction techniques, based e.g. on proper orthogonal decomposition (POD) of a set of snapshots \cite{sirovich1987turbulence,quarteroni2015reduced,hesthaven2016certified,guo2019data}, on fully connected auto-encoders, or on convolutional auto-encoders \cite{brunton2020machine,lee2020model,maulik2021reduced,fresca2021comprehensive,vlachas2022multiscale,oommen2022learning}.
The underlying assumption is that the dynamics can be represented by a limited number of state variables, called latent variables, whose time evolution is learned either through NNs with recurrent structure (such as RNNs \cite{liu2022hierarchical}, LSTMs \cite{maulik2021reduced,vlachas2022multiscale} or ODE-Nets \cite{chen2018neural}), dynamic mode decomposition \cite{brunton2020machine}, SINDy \cite{brunton2016discovering,rudy2017data}, fully-connected NNs (FCNNs) \cite{fresca2021comprehensive}, or DeepONets \cite{oommen2022learning}.

When a high-fidelity model is available, there are also techniques for building reduced-order models by exploiting knowledge of the equations \cite{prud2002reliable,benner2005dimension,antoulas2005approximation,benner2015survey,MOR2021,hesthaven2022reduced,bruna2022neural}.
These latter methods are however intrusive, unlike the formers, which learn a model in a data-driven manner using only a dataset of input-output pairs. 
Intrusive techniques are typically based on projecting the high-fidelity model into a low-dimensional space, obtained by POD or by greedy algorithms. 
In the case of nonlinear models, however, such techniques require special arrangements, such as the (discrete) empirical interpolation method \cite{barrault2004empirical,canuto2009posteriori,chaturantabut2010nonlinear}, but this entails a difficult trade-off between accuracy and computational cost.
Furthermore, many problems feature a slow decay Kolmogorov $n$-width, an index of the amenability of the solution manifold to be approximated by an $n$-dimensional linear subspace \cite{binev2011convergence,buffa2012priori}.
In many cases of interest, such as advection-dominated problems or high Reynolds number flow equations, POD-based methods achieve reasonable accuracy only for high values of $n$ \cite{lee2020model}.
This limits their use in practical applications.

In this paper, we present a novel family of NNs, called Latent Dynamics Networks (LDNets), that can effectively learn, in a data-driven manner, the temporal dynamics of space-dependent fields and predict their evolution for unseen time-dependent input signals and unseen scalar parameters.
LDNets automatically discover a compact encoding of the system state in terms of (typically a few) latent scalar variables.
Remarkably, the latent representation is learned without the need of using an auto-encoder to explicitly compress a high-dimensional discretization of the system state.
Furthermore, LDNets are based on an intrinsically space-dependent reconstruction of the output fields.
Indeed, instead of yielding a discrete representation of the fields (e.g. point values on a spatial mesh), LDNets are able to generate output fields defined in any point of space, in a \textit{meshless} manner.
As a consequence, the (typically high-dimensional) discrete representation of the output is never explicitly constructed.
These features make the training of LDNets extremely lightweight, and boost their generalization ability even in the presence of few training samples and even in time-extrapolation regimes, that is for longer time horizons than those seen during training.

	\section{Methods}\label{sec:PP:methods}

We introduce the notation used throughout this paper and we present the proposed method.

\subsection{Notation}

We denote by $\outputField \colon \spaceDomain \times [0, \timeMax] \to \mathbb{R}^{\NUMoutputField}$ an output field we aim to predict, where $\spaceDomain \subset \mathbb{R}^\NUMspacevar$ is the space domain and $\timeMax > 0$ is the final time.
The evolution of $\outputField$ is driven by the input $\inputSign \colon [0, \timeMax] \to \mathbb{R}^{\NUMinputSign}$, that is, a set of time-dependent signals or constant parameters.
Our goal is to unveil, starting from data, the laws underlying the dependence of $\outputField$ on $\inputSign$.

More precisely, we denote input signals as $\inputSign \colon [0, \timeMax] \to \SPACEinputSign$, taking values in the set $\SPACEinputSign \subseteq \mathbb{R}^{\NUMinputSign}$, and we denote by $\SETinputSign
\subseteq \{\inputSign \colon [0, \timeMax] \to \SPACEinputSign \}$ the set of admissible input signals. 
Then, we denote by $\outputField \colon \spaceDomain \times [0, \timeMax] \to \SPACEoutputField$ the output (space-time dependent) field, with values in $\SPACEoutputField \subseteq \mathbb{R}^{\NUMoutputField}$. 
Finally, we denote by $\SEToutputField
\subseteq \{\outputField \colon \spaceDomain \times [0, \timeMax] \to \SPACEoutputField \}$ the space of possible outputs.
We assume that the map $\inputSign \mapsto \outputField$ is well defined (i.e. the output $\outputField$ is unambiguously determined by the input $\inputSign$) and it is consistent with the arrow of time (i.e. $\outputField(\spacevar,\timevar)$ depends on $\inputSign(s)|_{s \in [0,\timevar]}$ but not on $\inputSign(s)|_{s \in (\timevar,\timeMax]}$).

An important (albeit not exclusive) example is the case when the dynamics we aim to learn underlies a differential model, that is when the map $\inputSign \mapsto \outputField$ is defined as the composition of an observation operator and the solution map $\inputSign \mapsto \stateFOM$ of a partial differential equation (PDE) in the form of
\begin{equation} \label{eqn:FOM}
    \left\{
    \begin{aligned}
        \partial_\timevar \stateFOM(\spacevar,\timevar) 
        &= 
        \FOMrhs(
            \stateFOM(\spacevar,\timevar), 
            \inputSign(\timevar)
            )
        & & \text{in $\spaceDomain \times (0, \timeMax]$}
        \\
        \outputField(\spacevar,\timevar) 
        &= 
        \FOMobs(
            \stateFOM(\spacevar,\timevar), 
            \spacevar
            )
        & & \text{in $\spaceDomain \times (0, \timeMax]$}
        \\
        \stateFOM(\spacevar,0)  
        &= \stateFOM_0(\spacevar)
        & & \text{in $\spaceDomain$,}
    \end{aligned}
    \right.
\end{equation}
where $\stateFOM \in \SETstateFOM \subseteq \{\stateFOM \colon \spaceDomain \times [0, \timeMax] \to \SPACEstateFOM \}$, with $\SPACEstateFOM \subseteq \mathbb{R}^{\NUMstateFOM}$, is the state variable (typically, $\SETstateFOM$ is a Sobolev space). Here, $\stateFOM_0 \colon \spaceDomain \to \SPACEstateFOM$ is the initial state, $\FOMrhs$ is a differential operator, and $\FOMobs$ is the observation operator.
In particular, LDNets can be also used for the model-order reduction of \eqref{eqn:FOM}, by passing through data generated via a numerical approximation of \eqref{eqn:FOM}, e.g. by the Finite Element method, called full-order model (FOM).
Meaningful examples are provided in Sec.~\ref{sec:PP:results}.
Nonetheless, in this paper neither knowledge nor even the existence of a model such as \eqref{eqn:FOM} is required: the training of an LDNet only requires input-output pairs.

\begin{remark}
    The case when the output field $\outputField$ is determined not only by some time-dependent inputs $\inputSign$, but also by some inputs that are time-independent (typically called \textit{parameters}) is a special case of the one considered here.
    Still, to keep the notation compact, we will use the same symbol $\inputSign$ to collectively denote time-dependent inputs (i.e. signals) and time-constant inputs (i.e. parameters).
\end{remark}

\begin{remark}
    The FOM \eqref{eqn:FOM} is an autonomous system.
    The non-autonomous case can be recovered as a special case by setting $\inputSign(\timevar) \cdot \mathbf{e}_k= \timevar$ for some $k$, where $\mathbf{e}_k$ is the $k$-th element of the canonical base of $\mathbb{R}^{\NUMinputSign}$.
\end{remark}

\subsection{Training data}

The training data are collected by considering a finite number of realizations of the map $\inputSign \mapsto \outputField$, each one referred to as a \textit{training sample}.
For each training sample $i \in \trainSamples$, we collect the following discrete observations:
\begin{itemize}
    \item $\inputSign_i(\OBStimevar)$, for $\OBStimevar \in \trainTIMESinputSign{i}$;
    \item $\outputField_i(\OBSspacevar, \OBStimevar)$, for $\OBStimevar \in \trainTIMESoutputField{i}$, $\OBSspacevar \in \trainPOINTSoutputField{i}{\OBStimevar}$;
\end{itemize}
where $\trainTIMESinputSign{i} \subset [0,\timeMax]$, $\trainTIMESoutputField{i} \subset [0,\timeMax]$ and $\trainPOINTSoutputField{i}{\OBStimevar} \subset \spaceDomain$ are discrete sets of observations.
We remark that the observation times and points can be either shared among samples (i.e. $\trainTIMESinputSign{i} \equiv \trainTIMESinputSign{}$, $\trainTIMESoutputField{i} \equiv \trainTIMESoutputField{}$ and $\trainPOINTSoutputField{i}{\OBStimevar} \equiv \trainPOINTSoutputField{}{}$ for any $i$ and for any $\OBStimevar$) or be different from one sample to another.

Our goal is to learn the map $\inputSign \mapsto \outputField$, that is to infer the output $\outputField(\spacevar, \timevar)$ corresponding to inputs $\inputSign(\timevar)$ outside the training set.

\subsection{LDNets}

\begin{figure}
    \centering
    \includegraphics[width=0.8\textwidth]{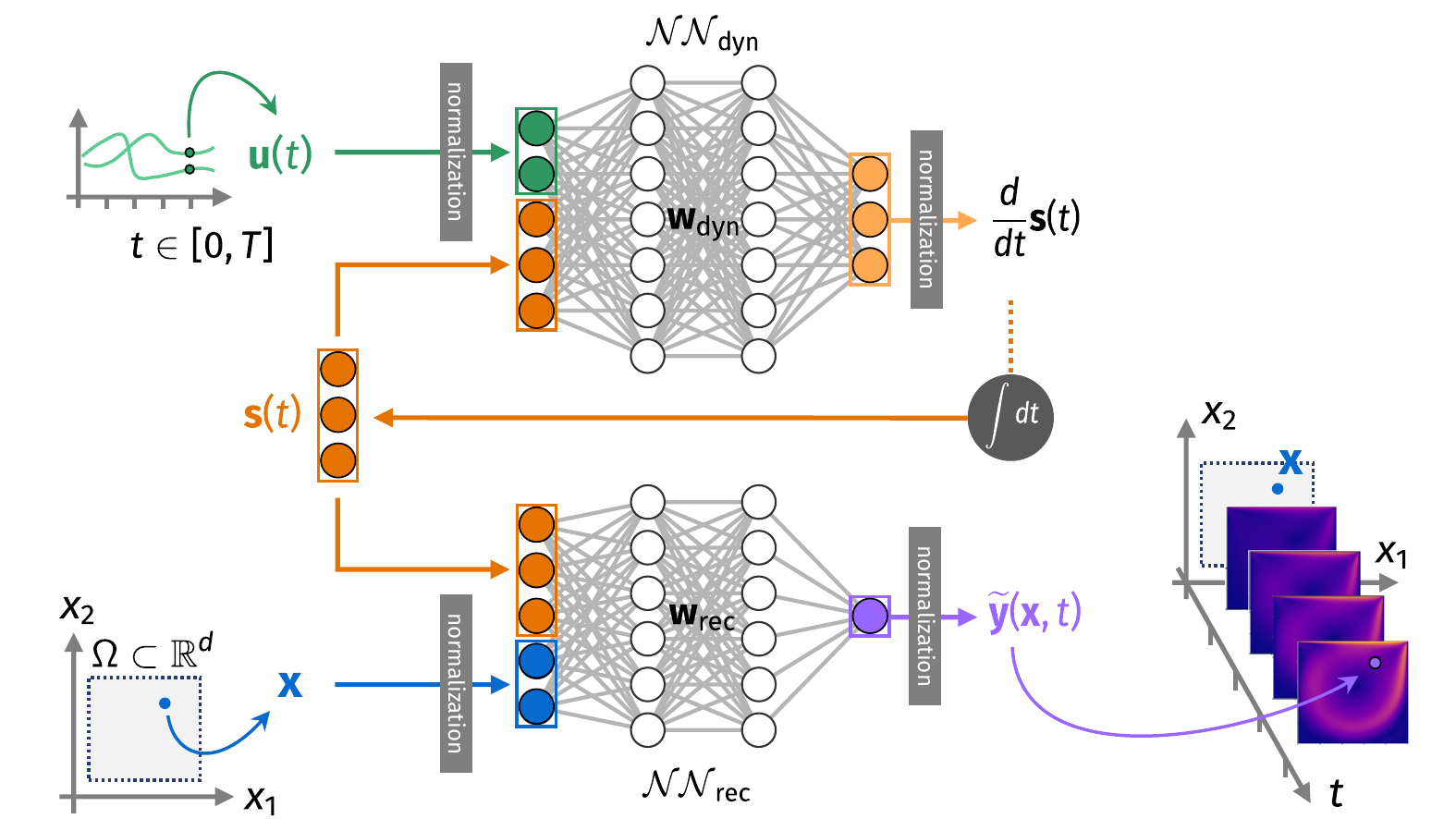}
    \caption{
        \textbf{LDNet architecture.}
        The network $\ROMrhs$ receives the input $\inputSign(\timevar)$ and the latent state $\stateROM(\timevar)$ and returns the time derivative of the latent state, thus defining its dynamics.
        The network $\ROMobs$, instead, is evaluated only when an estimate of the output field $\outputField$ is sought.
        More precisely, an approximation of $\outputField(\spacevar,\timevar)$ is recovered by giving as an input to $\ROMobs$ the latent state at time $t$ and the query space coordinate $\spacevar \in \spaceDomain$.
        In general, the reconstruction network $\ROMobs$ might take as an input $\inputSign(\timevar)$ as well (see e.g. Sec.~\ref{sec:PP:results}, Test Case 2); for simplicity, in the figure we represent the special case when $\ROMobs$ does not depend on $\inputSign(\timevar)$.
        }
    \label{fig:architecture}
\end{figure}

An LDNet consists of two sub-networks, $\ROMrhs$ and $\ROMobs$, that is two FCNNs with trainable parameters $\paramsROMdyn$ and $\paramsROMobs$, respectively (see Fig.~\ref{fig:architecture}).
The first NN, namely $\ROMrhs$, evolves the dynamics of the latent variables $\stateROM(\timevar) \in \mathbb{R}^{\NUMstateROM}$ according to the differential equation
\begin{equation} \label{eqn:LDNet_dynamics}
    \dot{\stateROM}(\timevar) 
        = 
        \ROMrhs(
            \stateROM(\timevar), 
            \inputSign(\timevar)
            ; \paramsROMdyn
            )
        \qquad
        \text{in $(0, \timeMax]$}
\end{equation}
with a prescribed arbitrary initial condition (in our numerical test, we set $\stateROM(0) = \mathbf{0}$).
The inputs of $\ROMrhs$ are the latent states $\stateROM(\timevar)$ and the input signal $\inputSign(\timevar)$ at the current time $t$.
Instead, the second NN, $\ROMobs$, is used to reconstruct $\APPoutputField$, an approximation of the output field $\outputField$ at any time $\timevar \in [0, \timeMax]$ and at any query point $\spacevar \in \spaceDomain$:
\begin{equation} \label{eqn:LDNet_reconstructions}
    \APPoutputField(\spacevar,\timevar) 
        = 
        \ROMobs(
            \stateROM(\timevar), 
            \inputSign(\timevar),
            \spacevar
            ; \paramsROMobs
            )
        \qquad
        \text{in $\spaceDomain \times (0, \timeMax]$}
\end{equation}
We remark that the reconstruction network $\ROMobs$ is independently queried for every point $\spacevar \in \spaceDomain$ for which the solution is sought.

Hence, the LDNet defines a map from a time-dependent input signal $\inputSign \in \SETinputSign$ to a space-time dependent field $\APPoutputField \in \SEToutputField$ through the solution of the following system of ordinary differential equations (ODEs):
\begin{equation}\label{eqn:ROM}
    \left\{
    \begin{aligned}
        \dot{\stateROM}(\timevar) 
        &= 
        \ROMrhs(
            \stateROM(\timevar), 
            \inputSign(\timevar)
            ; \paramsROMdyn
            )
        & & \text{in $(0, \timeMax]$}
        \\
        \stateROM(0)  
        &= \mathbf{0}
        \\
        \APPoutputField(\spacevar,\timevar) 
        &= 
        \ROMobs(
            \stateROM(\timevar), 
            \inputSign(\timevar),
            \spacevar
            ; \paramsROMobs
            )
        & & \text{for $\spacevar \in \spaceDomain$ and $\timevar \in [0, \timeMax]$,}
    \end{aligned}
    \right.
\end{equation}
where $\stateROM(\timevar) \in \mathbb{R}^{\NUMstateROM}$ is the vector of latent states.
The number of latent states $\NUMstateROM$ is set by the user, and should be regarded as an hyperparameter.
We remark that, thanks to the hidden nature of $\stateROM(\timevar)$, we can assume without loss of generality the initial condition $\stateROM(0) = \mathbf{0}$ (see \cite{regazzoni2019modellearning} for a discussion on this topic in a similar framework).
In this work, we always consider hyperbolic tangent ($\tanh$) activation functions.

\begin{remark}
The formulation \eqref{eqn:ROM} is the most general one. 
A special case is the one where $\ROMobs$ does not depend on $\inputSign(\timevar)$.
Whether or not to include the latter dependency in $\ROMobs$ is an architectural choice that shall be regarded as a hyperparameter, possibly subject to selection via cross-validation.
In many cases, however, the choice can be driven by the \textit{physics} of the underlying process.
Specifically, we will leave an explicit dependency whenever the output $\outputField(\spacevar,\timevar)$ depends on the input $\inputSign(\timevar)$ instantaneously.
The case where the dependency is neglected is the one that we will mostly consider in our test cases, expect for Test Case 2, in which we will allow $\ROMobs$ to depend on $\inputSign(\timevar)$ in a direct way.
\end{remark}

In practice, the ODE system \eqref{eqn:ROM} is discretized by a suitable numerical method.
In this work, we employ a Forward Euler scheme with a uniform time step size $\Dt$, but other schemes could be considered as well (e.g. time-adaptive Runge-Kutta schemes \cite{quarteroni2008numerical}).
In case the observation times $\trainTIMESinputSign{i}$ do not coincide with the discrete times $k \Dt$, for $k = 1, \dots$, we perform a re-sampling of $\inputSign$ through a piecewise linear interpolation.
Similarly, to evaluate the predicted output $\APPoutputField$ in correspondence of the observation times $\OBStimevar \in \trainTIMESoutputField{i}$, we interpolate the discrete solution of $\stateROM(\timevar)$ at the time instants $\OBStimevar$.

We denote with the symbol $\ROMdyn$ (to evoke its recurrent nature) the operator mapping the time series of inputs $\VECinputSign$ associated with a given sample $i$ to the latent state $\stateROM_i$ evolution.
More precisely, we have, for any sample $i$ and at any time $\timevar \in [0,\timeMax]$:
\begin{equation*}
    \stateROM_i(\timevar) = \ROMdyn(\VECinputSign, \timevar; \paramsROMdyn)
\end{equation*}
With this notation, the LDNet output $\APPoutputField_i(\spacevar,\timevar)$ is the result of the composition of $\ROMobs$ with $\ROMdyn$:
\begin{equation}\label{eqn:LDNet_output}
    \APPoutputField_i(\spacevar,\timevar) = \ROMobs(\ROMdyn(\VECinputSign, \timevar; \paramsROMdyn), \inputSign_i(\timevar), \spacevar; \paramsROMobs).
\end{equation}
To train the LDNet, we define the loss function:
\begin{equation*}
    \loss(\paramsROMdyn, \paramsROMobs) = 
    \avsum_{i \in \trainSamples}
    \avsum_{\OBStimevar \in \trainTIMESoutputField{i}}
    \avsum_{\OBSspacevar \in \trainPOINTSoutputField{i}{\OBStimevar}}
    \error(\APPoutputField_i(\OBSspacevar,\OBStimevar), \outputField_i(\OBSspacevar,\OBStimevar))
    +
    \weightREGrhs \lossREG(\paramsROMdyn)
    +
    \weightREGobs \lossREG(\paramsROMobs),
\end{equation*}
where the symbol $\avsum$ denotes the average operator (that is the sum over a set divided by the cardinality of the set), and where $\APPoutputField_i(\OBSspacevar,\OBStimevar)$ are the outputs of the LDNet associated with the trainable parameters $\paramsROMdyn$ and $\paramsROMobs$ as defined in \eqref{eqn:LDNet_output}. The discrepancy metric $\error$ is typically defined as 
\begin{equation} \label{eqn:discrepancy_metric}
    \error(\APPoutputField, \outputField)
    =
    \frac{\left\|
    \APPoutputField  - \outputField
    \right\|^2}{\NORMoutputField^2}
\end{equation}
with $\NORMoutputField$ being a normalization factor defined from case to case and where $\| \cdot \|$ denotes the euclidean norm.
The first term of $\loss$ represents therefore the normalized mean square error between observations and LDNet predictions.
Moreover, to mitigate overfitting, suitable regularization terms on the NN weights could be introduced, with weighting factors $\weightREGrhs$ and $\weightREGobs$.
In this work, we define $\lossREG$ as the mean of the squares of the NN weights (yielding the so-called $L^2$-regularization or Tikhonov regularization).

\begin{remark}
    The quadratic discrepancy metric \eqref{eqn:discrepancy_metric}, while being the most natural choice, is not the unique one.
    For instance, it can be replaced by goal-oriented metrics (an example is given in Test Case 2).
\end{remark}

Training an LDNet consists in employing suitable optimization methods to approximate the solution of the following non-convex minimization problem:
\begin{equation*}
    (\paramsROMdyn^*, \paramsROMobs^*) = \underset{\paramsROMdyn, \paramsROMobs}{\operatorname{argmin}} \loss(\paramsROMdyn, \paramsROMobs).
\end{equation*}
The two NNs are simultaneously trained. Thanks to the simultaneous end-to-end training of the two NNs, the latent space is discovered at the same time as learning the dynamics of the system.
This generalizes the approach presented in \cite{regazzoni2019modellearning} for the case of time signals as outputs.

\subsection{Normalization layers} \label{sec:methods:normalization}

In order to facilitate training, we normalize the inputs and the outputs of the NNs.
Specifically:
\begin{itemize}
    \item 
    We normalize the signals $\inputSign$, the output fields $\outputField$ and the space variables $\spacevar$, so that each entry approximately spans the interval $[-1, 1]$. 
    We normalize each entry independently of the others. 
    More precisely we normalize each scalar variable $\variable$ through the affine transformation
    $\ADIMvariable = (\variable - \variable_0) / \WIDTHvariable$
    where $\variable_0$ is a reference value and $\WIDTHvariable$ is a reference width.
    To define $\variable_0$ and $\WIDTHvariable$, we follow two different strategies.
    \begin{enumerate}
        \item If the variable takes values in a bounded interval $[\MINvariable, \MAXvariable]$, we set 
        \begin{equation*}
            \begin{split}
                \CENTERvariable &= (\MINvariable + \MAXvariable)/2, \\
                \WIDTHvariable  &= (\MAXvariable - \MINvariable)/2.
            \end{split}
        \end{equation*}
        \item If the variable is sampled from a distribution with unbounded support (e.g., when $\variable$ is normally distributed), we set $\CENTERvariable$ equal to the sample mean and $\WIDTHvariable$ equal to three times the sample standard deviation.
    \end{enumerate}

    \item
    We also normalize the time variable, by dividing the time steps by a characteristic time scale $\DtREF$. 
    The normalization constant $\DtREF$ impacts the output of $\ROMrhs$, that is dimensionally proportional to the inverse of time.
    Since finding a good value for $\DtREF$ is in general not straightforward, we typically consider it as a hyperparameter, tuned through a suitable automatic algorithm (see Section.~\ref{sec:SI:methods:hypertuning}).

    \item 
    We do not normalize the latent states $\stateROM$, since their distribution is not known before training.
    Indeed, when hyperparameters are well tuned, the training algorithm tends to generate models that produce latent states with approximately normalized values.
\end{itemize}
In practice, normalization can be achieved either by modifying the training data accordingly, or by embedding the two NNs between two normalization layers (namely, one input layer and one output layer) each. 
Formally, the second approach consists in defining $\ROMrhs$ and $\ROMobs$ as follows, where we $\adimROMrhs$ and $\adimROMobs$ are two FCNNs:
\begin{equation*}
    \begin{split}
    \ROMrhs(
        \stateROM, 
        \inputSign
        ; \paramsROMdyn
        )
        &=    
        \DtREF^{-1}
    \adimROMrhs(
        \stateROM, 
        (\inputSign - \CENTERinputSign) \oslash \WIDTHinputSign
        ; \paramsROMdyn
        )
    \\
    \ROMobs(
        \stateROM, 
        \inputSign,
        \spacevar
        ; \paramsROMobs
        )
        &=
        \CENTERoutputField +  \WIDTHoutputField \odot
        \adimROMobs(
            \stateROM, 
            (\inputSign - \CENTERinputSign) \oslash \WIDTHinputSign,
            (\spacevar - \CENTERspacevar) \oslash \WIDTHspacevar
            ; \paramsROMobs
            )
        \\
    \end{split}
\end{equation*}
where $\odot$ and $\oslash$ denote the Hadamard (i.e. element-wise) product and division, respectively.

\subsection{Imposing a-priori physical knowledge}

The architecture of LDNets reflects certain features of the physics they are meant to capture. 
With respect to the space variable, the representation is continuous, unlike methods that reconstruct a discretized solution thus losing the correspondence between neighboring points. 
With respect to the time variable, the dynamics is driven by a system of differential equations which makes LDNets consistent with the arrow of time (i.e., with the causality principle \cite{regazzoni2019modellearning}).
These features make it natural to introduce a-priori physical knowledge in the construction and training of LDNets.
In this regard, we distinguish between weak imposition and strong imposition.

Weak imposition consists of introducing physics-informed terms \cite{raissi2019physics} into the loss function, aimed at promoting solutions that satisfy certain requirements (such as irrotationality of a velocity field, to make an example). 
In this paper we do not show examples in this regard, but simply highlight that the continuous representation of the output field used by LDNets makes the introduction of such terms very straightforward through the use of automatic differentiation.

Strong imposition, on the other hand, consists of modifying the architecture of the LDNet components in order to obtain models that automatically satisfy certain properties \cite{as2022mechanics,linka2023new}. In what follows, we provide two examples of how this can be applied to ensure both temporal (acting on $\ROMrhs$) and spatial (acting on $\ROMobs$) properties.

\subsubsection{Equilibrium configuration imposition} \label{sec:methods:equilibrium}

In many real-life applications, data are collected starting from an equilibrium configuration.
This entails that the initial state should be an equilibrium for the latent dynamics as well, in virtue of the interpretation of $\stateROM$ as a compact encoding of the full-order system state.
Therefore, we define the right-hand side of the latent state evolution equation as follows, where $\adimROMrhs$ is a trainable FCNN and where $\EQinputSign \in \SPACEinputSign$ is the input at equilibrium:
\begin{equation*}
    \ROMrhs(
        \stateROM, 
        \inputSign
        ; \paramsROMdyn
        )
        =    
    \adimROMrhs(
        \stateROM, 
        \inputSign
        ; \paramsROMdyn
        )
        -   
    \adimROMrhs(
        \mathbf{0}, 
        \EQinputSign
        ; \paramsROMdyn
        )
\end{equation*}
As a consequence, the initial state $\stateROM = \mathbf{0}$ of the model is an equilibrium for any choice of the trainable parameters $\paramsROMdyn$.

\subsubsection{Prescribed solution in subsets of the domain (e.g. Dirichlet boundary conditions)}

The evolution of the output field is often unknown except on a subset of the domain $\spaceDomain$, such as for example a portion $\Gamma_D$ of its boundary $\partial\spaceDomain$.
This happens, e.g., when there is a FOM that features a Dirichlet boundary condition like
\begin{equation}\label{eqn:dirichlet_BC}
    \outputField(\spacevar, \timevar) = \DirichletDatum(\spacevar) \quad \text{on $\Gamma_D$}.
\end{equation}
In this case, the solution is constrained to satisfy \eqref{eqn:dirichlet_BC} by defining $\ROMobs$ as
\begin{equation*}
    \ROMobs(
        \stateROM, 
        \inputSign,
        \spacevar
        ; \paramsROMobs
        )
        =
        \DirichletLifting(\spacevar) + 
    \adimROMobs(
        \stateROM, 
        \inputSign,
        \spacevar
        ; \paramsROMobs
        )
        \DirichletMask(\spacevar)
\end{equation*}
where $\adimROMobs$ is a trainable FCNN, $\DirichletLifting$ is the lifting of the boundary datum, that is an extension of $\DirichletDatum$ to the whole domain $\spaceDomain$, and $\DirichletMask\colon\Omega\to\mathbb{R}$ is a mask, that is a smooth function such that $\DirichletMask(\spacevar) = 0$ if and only if $\spacevar \in \Gamma_D$. See \cite{regazzoni2022usmnets} for further details and \cite{berrone2022enforcing} for a general approach to construct the mask $\DirichletMask$ based on approximate distance functions.

\subsection{Error metrics}

To evaluate the generalization accuracy of a trained LDNet, we test it on unseen data, that is on samples belonging to a test set denoted by $\testSamples$. 
Specifically, we employ two metrics.

\vspace{0.3cm}
\noindent\textbf{Normalized root mean square error (NRMSE)} is obtained as the square root of the mean of the squares of the normalized errors obtained on the testing set:

\begin{equation*}
    \mathrm{NRMSE} = 
    \sqrt{
    \avsum_{i \in \testSamples}
    \avsum_{\OBStimevar \in \trainTIMESoutputField{i}}
    \avsum_{\OBSspacevar \in \trainPOINTSoutputField{i}{\OBStimevar}}
    \frac{\left\|
    \APPoutputField_i(\OBSspacevar,\OBStimevar)  - \outputField_i(\OBSspacevar,\OBStimevar) 
    \right\|^2}{\NORMoutputField^2}
    }.
\end{equation*}

\vspace{0.3cm}
\noindent\textbf{Pearson dissimilarity ($1 - \rho$)} is defined from the Pearson correlation coefficient $\rho$:
\begin{equation*}
    \rho = 
    \frac{
        \displaystyle\sum_{i \in \testSamples}
        \displaystyle\sum_{\OBStimevar \in \trainTIMESoutputField{i}}
        \displaystyle\sum_{\OBSspacevar \in \trainPOINTSoutputField{i}{\OBStimevar}}
        (\APPoutputField_i(\OBSspacevar,\OBStimevar) - \overline{\APPoutputField}) \cdot
        (   \outputField_i(\OBSspacevar,\OBStimevar) - \overline{   \outputField})
    }{
    \sqrt{
        \displaystyle\sum_{i \in \testSamples}
        \displaystyle\sum_{\OBStimevar \in \trainTIMESoutputField{i}}
        \displaystyle\sum_{\OBSspacevar \in \trainPOINTSoutputField{i}{\OBStimevar}}
        \|\APPoutputField_i(\OBSspacevar,\OBStimevar) - \overline{\APPoutputField}\|^2
        \displaystyle\sum_{i \in \testSamples}
        \displaystyle\sum_{\OBStimevar \in \trainTIMESoutputField{i}}
        \displaystyle\sum_{\OBSspacevar \in \trainPOINTSoutputField{i}{\OBStimevar}}
        \|   \outputField_i(\OBSspacevar,\OBStimevar) - \overline{   \outputField}\|^2
    }
    }
\end{equation*}
where we denote the average outputs as
\begin{equation*}
    \begin{split}
        \overline{   \outputField} &= 
        \avsum_{i \in \testSamples}
        \avsum_{\OBStimevar \in \trainTIMESoutputField{i}}
        \avsum_{\OBSspacevar \in \trainPOINTSoutputField{i}{\OBStimevar}}
           \outputField_i(\OBSspacevar,\OBStimevar)
        \\
        \overline{\APPoutputField} &= 
        \avsum_{i \in \testSamples}
        \avsum_{\OBStimevar \in \trainTIMESoutputField{i}}
        \avsum_{\OBSspacevar \in \trainPOINTSoutputField{i}{\OBStimevar}}
        \APPoutputField_i(\OBSspacevar,\OBStimevar)
    \end{split}
\end{equation*}
We remark that both metrics (RMSE and $1 - \rho$) are robust with respect to multiplicative rescaling of the outputs. For both of them, the smaller the value of the metric, the higher the accuracy of the predictions.

\subsection{Training algorithm}

To train the LDNet, we employ a two stage strategy.
First, we perform a limited number of epochs (typically, a few hundreds) with the Adam optimizer \cite{kingma2014adam}, starting with a learning rate of $10^{-2}$.
Then, we switch to a second-order accurate optimizer, namely BFGS \cite{goodfellow2016deep}.
BFGS is more accurate than Adam, but more prone to get stuck in local minima, which is why it is useful to precede it with some Adam iterations, which provide a good initial guess.

To evaluate the gradient of the loss function with respect to the trainable parameters, we combine back-propagation-through-time for $\ROMdyn$ with back-propagation for $\ROMobs$ \cite{goodfellow2016deep}.
To initialize the parameters of the two NNs, we employ a Glorot uniform strategy for weights and zero values for the biases \cite{goodfellow2016deep}.

\subsection{Hyperparameters tuning algorithms} \label{sec:SI:methods:hypertuning}

The hyperparameters of the proposed method are the number of layers and neurons of $\ROMrhs$ and $\ROMobs$, the $L^2$ regularization weights $\weightREGrhs$ and $\weightREGobs$, the normalization time constant $\DtREF$ and, whenever necessary in the different test cases, the number of latent states $\NUMstateROM$.
To automatically tune them, we employ the Tree-structured Parzen Estimator (TPE) Bayesian algorithm \cite{bergstra2011algorithms,optuna2019}.
The hyperparameters search space is defined as an hypercube, with a log-uniform sampling.
We perform K-fold cross validation while monitoring the value of the discrepancy metric in Eq.~\eqref{eqn:discrepancy_metric}.
We also employ the Asynchronous Successive Halving (ASHA) scheduler to early terminate hyperparameters configurations that are either bad or not promising \cite{asha2020,hyperband2017}.

We simultaneously train multiple NNs associated to different hyperparameters settings on a supercomputer endowed with several CPUs via Message Passing Interface (MPI).
Each NN exploits Open Multi-Processing (OpenMP) for Hyper-Threading, which allows for a speed-up in the computationally-intensive tensorial operations involved during the training phase.
For the implementation, we rely on the Ray Python distributed framework \cite{ray2018}.
	\section{Results} \label{sec:PP:results}

\newcommand{\myrange}[2]{$#1$ -- $#2$}

We demonstrate the effectiveness of LDNets through several test cases. 
First, we consider a linear PDE model to analyze the ability of LDNets to extract a compact latent representation of models that are progressively less amenable to reduction.
Then, we consider the time-dependent version of a benchmark problem in fluid dynamics.
Finally, we compare LDNets with state-of-the-art methods in a challenging task, that is, learning the dynamics of the Aliev-Panfilov model \cite{aliev1996simple}, a highly non-linear excitation-propagation PDE model used in the field of cardiac electrophysiology modeling.
We focus on synthetically generated data obtained by numerical approximation of differential models, thus allowing us to test LDNet predictions against ground-truth results.

\subsection{Test Case 1: Advection-Diffusion-Reaction equation}

We consider the linear advection-diffusion-reaction (ADR) equation on the interval $\spaceDomain = (-1, 1)$:
\begin{equation} \label{eqn:FOM_ADR}
        \begin{aligned}
            &\frac{\partial \ADRstate(x,t)}{\partial t} 
            - \mu_1 \frac{\partial^2\ADRstate(x,t)}{\partial x^2}
            - \mu_2 \frac{\partial  \ADRstate(x,t)}{\partial x  }
            + \mu_3 \ADRstate(x,t) 
            = f(x,t) 
            \quad & & x \in (-1, 1), \, t \in (0, T].\\
        \end{aligned}
\end{equation}
This PDE is widely used, e.g., to describe the concentration $\ADRstate(x,t)$ of a substance dissolved in a channel \cite{quarteroni2008numerical}.
The constant parameters $\mu_1$, $\mu_2$ and $\mu_3$ respectively represent diffusion, advection and reaction coefficients, while the forcing term $f(x,t)$ is a prescribed external source, defined as $f(x,t) = A(t) \cos(2 \pi F(t) x - P(t))$, that is a sine wave with amplitude $A(t)$, frequency $F(t)$ and phase $P(t)$.
We consider an initial condition $\ADRstate(x, 0) = \ADRstate_0(x)$ and periodic boundary conditions.

To generate the training dataset, we employ a high-fidelity FFT-based solver on 101 equally spaced grid points, combined with an adaptive-time integration scheme for stiff problems \cite{brunton2022data,petzold1983automatic}.
Then, we subsample the time domain in 100 equally distributed intervals.
We challenge LDNets in predicting the space-time evolution of the target variable $\outputField(x,t) = \ADRstate(x,t)$ by considering three cases of increasing complexity (Test Cases 1a, 1b, 1c), in which the input $\inputSign$ is associated either to the parameters $\mu_1$, $\mu_2$ and $\mu_3$, or to the forcing term $f(x,t)$.
In all the test cases, we define $\NORMoutputField$ as the difference between maximum and minimum value taken by the output on the whole training set.
In all the cases presented below, unless otherwise stated, we set $\Dt = \num{5e-2}$.


\subsubsection{Sampling of inputs} \label{sec:SI:results:ADR:sampling}

In order to generate training and testing data, we need to sample the space of inputs.
This calls for a probability distribution on the latter space.
For the parameters ($\mu_1$, $\mu_2$ and $\mu_3$), we take for simplicity a uniform distribution on a suitable hypercube.
For the time-dependent signals ($A(t)$, $F(t)$ and $P(t)$), instead, probability distributions on function spaces are needed. 
For $A(t)$ and $P(t)$, we consider a Gaussian Process distribution \cite{rasmussen2003gaussian} with mean $\mu$ and with the following covariance kernel
\begin{equation*}
    K(t_1,t_2) = \sigma^2 \operatorname{exp}{\left[ - \frac{(t_1-t_2)^2}{2 \tau^2}\right]},
\end{equation*}
characterized by standard deviation $\sigma$ and characteristic time-scale $\tau$.
For what concerns $F(t)$ instead, in order to let it vary within a bounded set $(\fmin,\fmax)$, we define it as
\begin{equation*}
    F(t) = \frac{1}{2} \left[ \fmin + \fmax + (\fmax - \fmin) \tanh\left(\frac{3}{5} \gamma(t)\right)\right],
\end{equation*}
with $\gamma(t)$ sampled from a Gaussian Process with mean $\mu = 0$, standard deviation $\sigma = 1$ and prescribed characteristic time-scale $\tau$.
The values of $\mu$, $\sigma$ and $\tau$ are indicated below for each test case.

\subsubsection{Test Case 1a: finite latent dimension, constant parameters}

First, we consider $\ADRstate_0(x) = \cos(\pi x)$ and $f \equiv 0$.
We aim at predicting the evolution of $\ADRstate(x,t)$, depending on the constant parameters $\inputSign(\timevar) \equiv (\mu_1, \mu_2, \mu_3)$.
Due to the linearity of the equation, the solution is, at any time $t$, a sine wave with period 2, and can be thus unambiguously identified by two scalars (namely, the wave amplitude and phase, or equivalently, the real and imaginary part of the Fourier transform at frequency 0.5).
In other terms, the intrinsic dimension of the solution manifold is strictly equal to 2.
This provides therefore an ideal testbed for the capability of LDNets to recognize and learn a low-dimensional encoding of the system state from data.

To generate training and testing data, we employ a Monte Carlo sampling of the hypercube defined by the bounds $(\mu_1, \mu_2, \mu_3) \in [0, 0.05] \times [-0.1, 0.1] \times [0, 0.01]$.
We consider 100 training samples and 500 testing samples.
We select the hyperparameters according to the algorithm presented in Sec.~\ref{sec:SI:methods:hypertuning}.
The ranges of hyperparameter values used in the tuning algorithm are reported in Tab.~\ref{tab:ADR:hyperparameters}, in the row \textit{tuning}.
The selected values are instead reported in the row \textit{final}.

\begin{sidewaystable}
	\centering
	\begin{tabular}{ll|cccccc|cc}
		\toprule
        \multicolumn{2}{l|}{\textbf{Test cases}}  &\multicolumn{6}{c|}{\textbf{Hyperparameters}}                                                                        & \multicolumn{2}{c}{\textbf{Trainable parameters}} \\
                &                        &\multicolumn{2}{c}{$\ROMrhs$}        & \multicolumn{2}{c}{$\ROMobs$}   & $\DtREF$            & $\weightREGrhs$, $\weightREGobs$  & $\ROMrhs$ & $\ROMobs$ \\
		        &                        & layers         & neurons            & layers    & neurons             &                     &                                   & \# param. & \# param. \\ \midrule \multicolumn{2}{l}{\textbf{Test Case 1a}} & \multicolumn{8}{l}{}\\
        \multicolumn{2}{l|}{tuning      }& 2              & \myrange{3}{30}    & 2         & \myrange{3}{15}     & $[10^{-1}, 10^{1}]$ & 0                                 &           &          \\
        \multicolumn{2}{l|}{final       }& 2              & 9                  & 2         & 11                  & 0.5                 & 0                                 & 164       & 188      \vspace{.25cm} \\ \midrule \multicolumn{2}{l}{\textbf{Test Case 1b}} & \multicolumn{8}{l}{}\\
        \multicolumn{2}{l|}{tuning      }& 2              & \myrange{3}{30}    & 2         & \myrange{3}{15}     & $[10^{-1}, 10^{1}]$ & $[10^{-6}, 10^{-1}]$              &           &          \\
        \multicolumn{2}{l|}{final       }& 2              & 10                 & 2         & 7                   & 2.3                 & $10^{-5}$                         & 182       &  92      \vspace{.25cm} \\ \midrule \multicolumn{2}{l}{\textbf{Test Case 1c}} & \multicolumn{8}{l}{}\\
        \multicolumn{2}{l|}{tuning}      & \myrange{1}{3} & \myrange{5}{20}    & 1,2       & \myrange{4}{15}     & $[10^{-1}, 10^{1}]$ & $[10^{-5}, 10^{-1}]$              &           &          \\
        \multicolumn{2}{l|}{fine tuning }& 2              & twice $\ROMobs$    & 2         & \myrange{4}{9}      & 8                   & $[\num{2e-4},\num{2e-3}]$         &           &          \\ $\fmax$ & \multicolumn{1}{l}{$\NUMstateROM$} & \multicolumn{8}{l}{} \\
        0.5     & \multicolumn{1}{l|}{2} & 2              & 10                 & 2         & 5                   & 8                   & \num{1.95e-03}                    & 192       &  56       \\
        0.5     & \multicolumn{1}{l|}{3} & 2              & 16                 & 2         & 8                   & 8                   & \num{2.10e-04}                    & 435       & 121       \\
        0.5     & \multicolumn{1}{l|}{4} & 2              & 16                 & 2         & 8                   & 8                   & \num{2.10e-04}                    & 468       & 129       \\ \vspace{.25cm}
        0.5     & \multicolumn{1}{l|}{5} & 2              & 16                 & 2         & 8                   & 8                   & \num{2.00e-04}                    & 501       & 137       \\ 
        1       & \multicolumn{1}{l|}{2} & 2              & 12                 & 2         & 6                   & 8                   & \num{1.05e-03}                    & 254       &  73       \\
        1       & \multicolumn{1}{l|}{3} & 2              & 14                 & 2         & 7                   & 8                   & \num{4.20e-04}                    & 353       &  99       \\
        1       & \multicolumn{1}{l|}{4} & 2              & 16                 & 2         & 8                   & 8                   & \num{2.70e-04}                    & 468       & 129       \\  \vspace{.25cm}
        1       & \multicolumn{1}{l|}{5} & 2              & 16                 & 2         & 8                   & 8                   & \num{3.50e-04}                    & 501       & 137       \\
        2       & \multicolumn{1}{l|}{2} & 2              & 14                 & 2         & 7                   & 8                   & \num{6.60e-04}                    & 324       &  92       \\
        2       & \multicolumn{1}{l|}{3} & 2              & 16                 & 2         & 8                   & 8                   & \num{3.90e-04}                    & 435       & 121       \\
        2       & \multicolumn{1}{l|}{4} & 2              & 16                 & 2         & 8                   & 8                   & \num{2.20e-04}                    & 468       & 129       \\
        2       & \multicolumn{1}{l|}{5} & 2              & 16                 & 2         & 8                   & 8                   & \num{2.00e-04}                    & 501       & 137       \\
        \bottomrule
	\end{tabular}
	\caption{Test Case 1: hyperparameters ranges and selected values. 
    We remark that, in the fine tuning stage of test Case 1c, we select twice as many neurons for $\ROMrhs$ as for $\ROMobs$, in order to reduce the number of independent hyperparameters.    
    See text for details.}
	\label{tab:ADR:hyperparameters}
\end{sidewaystable}

In Tab.~\ref{tab:ADR_a:results} we report the training and testing accuracy metrics obtained by training an LDNet with $\NUMstateROM = 2$ latent variables.
In the table we show the accuracy achieved after 500, 5000 or 50000 epochs of BFGS (in all the cases, we first run 200 epochs of Adam).
The LDNet, trained on 100 samples, achieves an excellent accuracy when tested on 500 unseen samples.
Indeed, the NRMSE is $\num{1.88e-5}$ on the testing set, against a training NRMSE of $\num{1.81e-5}$.
Pearson dissimilarity is $\num{3.30e-9}$ on the testing set and $\num{3.00e-9}$ on the training set.
The very small differences in the accuracy metrics between training and testing sets provide evidence that the trained LDNet reproduces the FOM dynamics with great fidelity and without overfitting, that is with remarkably good generalization capabilities.

\begin{table}
	\centering
	\begin{tabular}{l|rrr}
		\toprule
		BFGS epochs & 500 & 5000 & 50000\\
		\midrule
        Training time & 8m 5s & 1h 5m & 10h 8m \\
		NRMSE\textsubscript{train}    & \num{1.02e-03} & \num{7.08e-05} & \num{1.81e-05} \\
		NRMSE\textsubscript{test}     & \num{1.19e-03} & \num{7.23e-05} & \num{1.88e-05} \\
		$1 - \rho_{\mathrm{train}}$   & \num{9.42e-06} & \num{4.58e-08} & \num{3.00e-09} \\
		$1 - \rho_{\mathrm{test} }$   & \num{1.33e-05} & \num{4.87e-08} & \num{3.30e-09} \\
		\bottomrule
	\end{tabular}
	\caption{Test Case 1a: training and test accuracy metrics for LDNets trained with an increasing number of BFGS training epochs (500, 5000 and 50000). Training time refer to a single-CPU standard laptop.}
	\label{tab:ADR_a:results}
\end{table}

\subsubsection{Test Case 1b: finite latent dimension, time-dependent inputs}

We now consider the case of time-dependent inputs, with a forcing term $f(x,t) = A(t) \cos(\pi x - P(t))$, where $\inputSign(t) = (A(t), P(t))$ represents an input signal that can vary in time within a bounded set.
We thus fix the values of the parameters to $\mu_1 = 0.05$, $\mu_2 = 0$ and $\mu_3 = 0.002$.
Similarly to Test Case 1a, the solution manifold has dimension 2 (thanks to the equation being linear and to the forcing term having constant frequency), but learning the dynamics becomes more challenging due to the presence of time-dependent inputs.
We let the amplitude $A(t)$ and the phase $P(t)$ vary in time, as described in Sec.~\ref{sec:SI:results:ADR:sampling}.
Specifically, for $A(t)$ we set $\mu = 2/5$, $\sigma = 2/15$ and $\tau = 1$, while for $P(t)$ we set $\mu = 0$, $\sigma = 4/3$ and $\tau = 1$.

\begin{figure}
    \includegraphics[width=\textwidth]{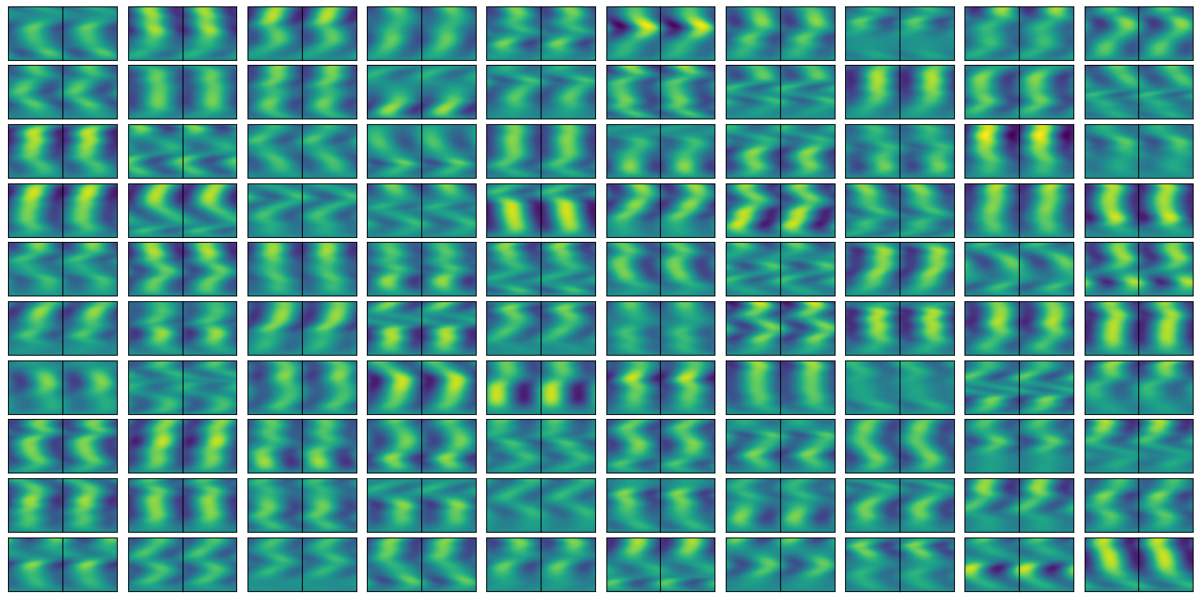}
    \caption{Test Case 1b. 100 testing samples, comparing the reference outputs $\outputField$ (left) with the LDNet predictions $\APPoutputField$ (right).
    For each sample, the horizontal axis refers to space, and the vertical axis refers to time.}
    \label{fig:ADR_am-ph_samples}
\end{figure}

First, we consider 100 training samples and we set $\Dt = 0.05$.
We tune the hyperparameters starting from the ranges indicated in Tab.~\ref{tab:ADR:hyperparameters}, and we select the optimal values reported in the same table.
The LDNets predictions $\APPoutputField$ for some test samples are displayed against reference outputs $\outputField$ in Fig.~\ref{fig:ADR_am-ph_samples}.

\begin{figure}
    \includegraphics[width=\textwidth]{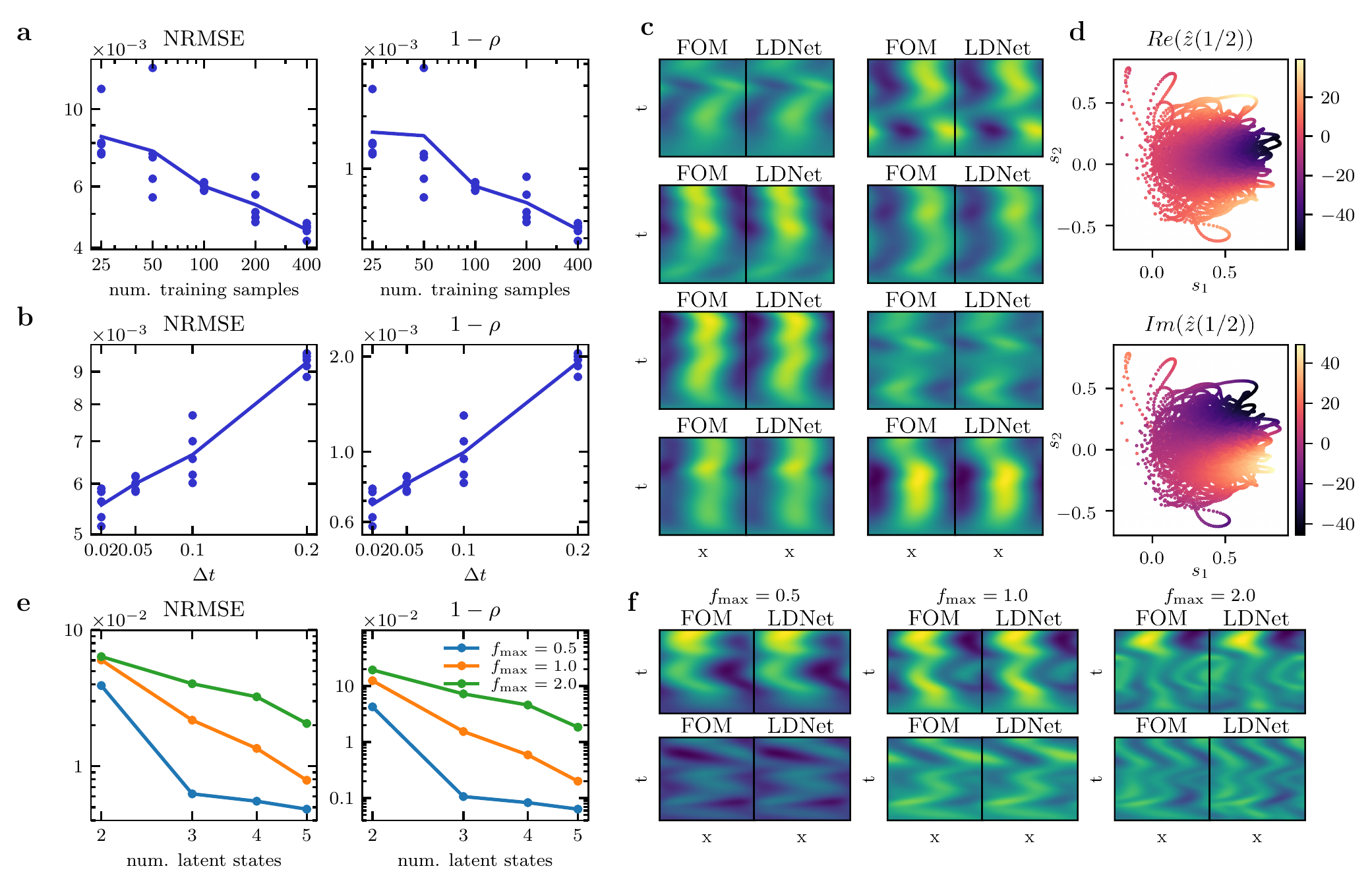}
    \caption{
        \textbf{Results of Test Case 1.}
        \figpt{a}-\figpt{b}: Testing accuracy of Test Case 1b, as a function of the number of training samples (with $\Delta t = 0.05$) and of $\Delta t$ (with 100 training samples). For each setting we run 5 training runs with random weights initialization. Each dot corresponds to a training run, while the solid line is the geometric mean.
        \figpt{c}: FOM against LDNet predictions on 8 testing samples for Test Case 1b. The abscissa corresponds to space and the ordinate to time.
        \figpt{d}: Mapping from the latent space trajectories and the Fourier space coefficients of the FOM solution for the testing samples of Test Case 1b.
        \figpt{e}: Testing accuracy of Test Case 1c as a function of the number of latent states and of the maximum input frequency $\fmax$.
        \figpt{f}: FOM against LDNet predictions on 2 testing samples for Test Case 1c, obtained by employing 5 latent states, for different maximum input frequencies (reported above the figure).
        }
    \label{fig:ADR_grid}
\end{figure}

Now, we perform two tests. 
First, by keeping $\Dt = 0.05$, we vary the number of training samples in the set $\{25, 50, 100, 200, 400\}$.
Then, we let $\Dt$ vary in the set $\{0.2, 0.1, 0.05, 0.02 \}$, by keeping the number of training samples equal to 100.
In both the cases, we do not vary the hyperparameters reported in Tab.~\ref{tab:ADR:hyperparameters}.
As desirable, the accuracy of predictions improves as the number of training samples increases and as the time discretization step size is reduced (Fig.~\ref{fig:ADR_grid}b).

Test Case 1b provides an ideal testbed to assess the capabilities of LDNets in discovering a compact representation of the FOM state.
The state of \eqref{eqn:FOM_ADR} indeed evolves on a two-dimensional manifold, being the state fully characterized, at any time $t$, by two scalars.
In other terms, provided that the forcing term is defined as in Test Case 1b, the model \eqref{eqn:FOM_ADR} has an intrinsic latent dimension equal to 2.
At any time $t$, in fact, $z(\cdot, t)$ is a sine wave with frequency 0.5.
Among the infinitely many equivalent parametrization, one is given by the Fourier transform of $z(\cdot, t)$ at frequency 0.5, that is determined by its real and imaginary part, respectively denoted by $Re(\hat{z}(0.5))$ and $Im(\hat{z}(0.5))$.

\begin{figure}
    \includegraphics[width=\textwidth]{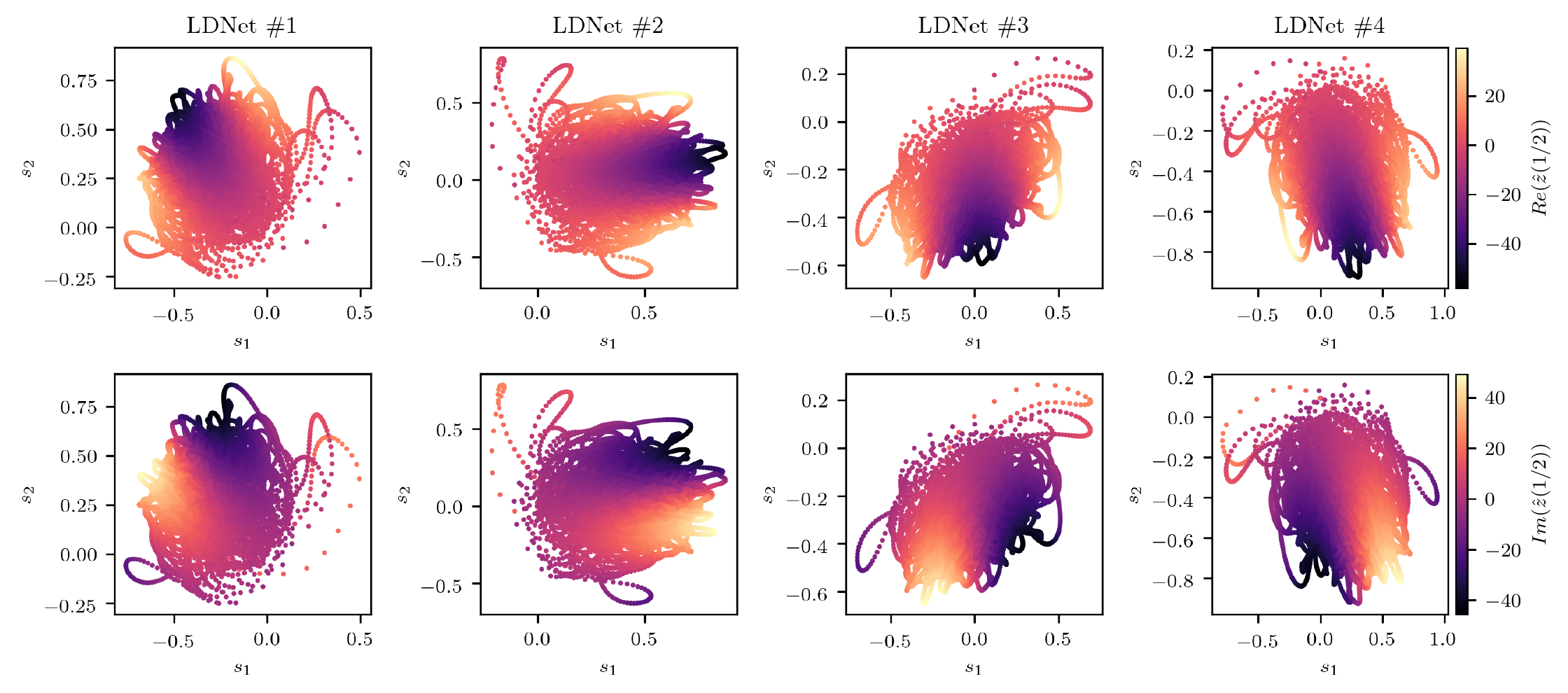}\\
    \caption{Test Case 1b. Trajectories in the latent space $(s_1, s_2)$ of 24 testing samples obtained with four different LDNets, by starting from as many different initial guesses for the trainable parameters (each of them corresponding to different columns).
    In the first (respectively, second) row, each point in the latent space is colored according to the corresponding value of $Re(\hat{z}(0.5))$ (respectively, $Im(\hat{z}(0.5))$).}
    \label{fig:ADR_phase_loops_multiple}
\end{figure}

Our results show that, during the training process of an LDNet, the algorithm discovers a compact representation of the solution field $z(\cdot, t)$, represented by the two latent variables $s_1(t)$ and $s_2(t)$.
We now investigate whether there is a relationship between the pairs $(s_1, s_2)$ and $(Re(\hat{z}(0.5)), Im(\hat{z}(0.5)))$.
With this goal, we train four different LDNets, starting from a different random initializations of the trainable parameters.
Then, we evaluate the trained LDNets on 24 test samples, and we collect the trajectories in the latent space $(s_1, s_2)$.
Finally, we plot these trajectories by displaying each point with a color that depends on the corresponding value of $Re(\hat{z}(0.5))$ computed from the reference solution, and we repeat the same procedure by considering the values of $Im(\hat{z}(0.5))$.
The results are shown in Fig.~\ref{fig:ADR_phase_loops_multiple}.
We notice that trajectories significantly differ among the four LDNets.
This is not surprising, as latent states are internal variables in the LDNet, hidden within the input-output relationship.
Nonetheless, a common pattern emerges in the connection between the latent states and the Fourier coefficients: each of the four trained LDNets underlies a well-defined relation between $(s_1, s_2)$ and $(Re(\hat{z}(0.5)), Im(\hat{z}(0.5)))$.
As a matter of fact, each LDNet discovers a different compact encoding for the FOM state, each of which underlying a relationship with the Fourier coefficients of the solution.
In other terms, despite not being explicitly instructed to do that, the LDNet \textit{discovers} an operator that is equivalent to the Fourier transform of the state.
At the same time, the reconstruction NN ($\ROMobs$) \textit{discovers} the inverse operator, as it is able to reconstruct the function $z(\cdot, t)$ from the two scalars $(s_1, s_2)$.
Remarkably, this is obtained in a fully data-driven manner, without explicitly using any prior Fourier-based feature extraction.

\subsubsection{Test Case 1c: infinite latent dimension}

Finally, we consider a forcing term $f(x,t) = A(t) \cos(2 \pi F(t) x - P(t))$, where $\inputSign(t) = (A(t), F(t), P(t))$ is the time-dependent input signal. The forcing frequency $F(t)$ varies within an interval $[0.25, \fmax]$. Hence, the solution manifold of \eqref{eqn:FOM_ADR} has a potentially infinite dimension, being $\ADRstate$ the superimposition of a continuum of frequencies. 
Still, the results show that LDNets are able to discover effective low-dimensional encodings of the state.
Specifically, to sample $F(t)$ we set $\tau = 1$, $\fmin = 0.25$ and $\fmax$ as indicated below; for $A(t)$ we set $\mu = 1$, $\sigma = 1/3$ and $\tau = 1$; for $P(t)$ we set $\mu = 0$, $\sigma = 4/3$ and $\tau = 1$.

\begin{figure}
    \includegraphics[width=\textwidth]{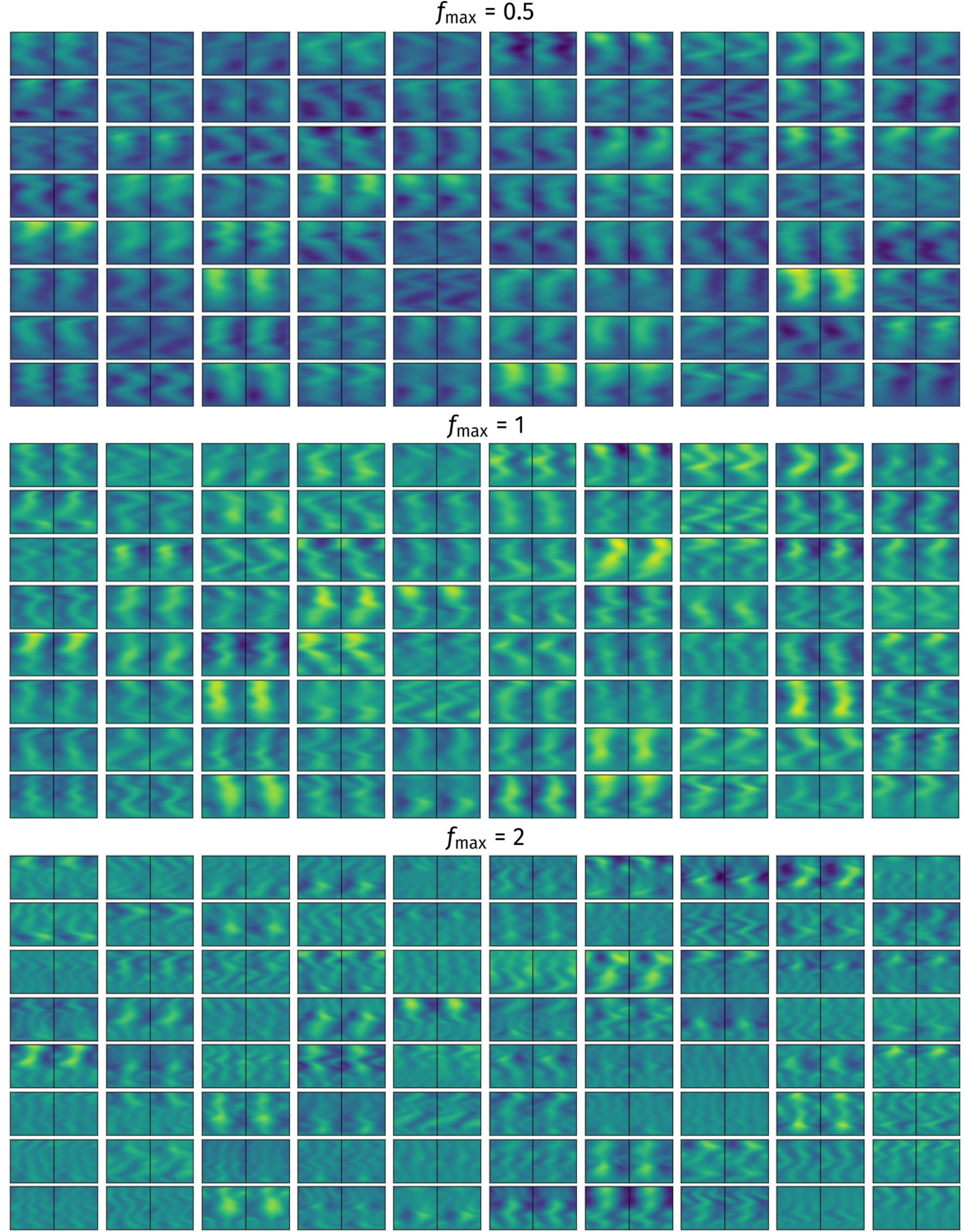}
    \caption{Test Case 1c. 80 testing samples for each $\fmax$ value considered in this work (namely 0.5, 1 and 2), comparing the reference outputs $\outputField$ (left) with the LDNet predictions $\APPoutputField$ (right), for $\NUMstateROM = 5$.
    For each sample, the horizontal axis refers to space, and the vertical axis refers to time.}
    \label{fig:ADR_am-ph-fq_samples}
\end{figure}

In this test case, we are interested in studying the impact of the number of latent states $\NUMstateROM$ on the LDNet accuracy, in three increasingly challenging cases, namely by setting $\fmax = 0.5$, $\fmax = 1$ and $\fmax = 2$ (see Fig.~\ref{fig:ADR_am-ph-fq_samples}).
In all the cases, we consider $\NUMstateROM \in \{2,3,4,5\}$.
For each combination of $\fmax$ and $\NUMstateROM$, we retune the hyperparameters, in order to compare the (in principle) best accuracy attainable.
With this purpose, we first run a preliminary hyperparameters tuning step, by considering a wide range of values (see Tab.~\ref{tab:ADR:hyperparameters}, row \textit{tuning}).
Then, we shrink the variability by selecting the most (generally with respect to different combinations) promising area of the hyperparameter space, and we perform a fine tuning for each combination of $\fmax$ and $\NUMstateROM$ independently (row \textit{fine tuning}).
The selected hyperparameters are listed in Tab.~\ref{tab:ADR:hyperparameters}.

First, we set $\fmax = 0.5$ and we train LDNets for increasing number of latent states, from 2 to 5. Remarkably, as the number of latent states increases, LDNets discover more effective encodings, that reflect in an increasing prediction accuracy (Fig.~\ref{fig:ADR_grid}e, blue line), with a saturation due to the relatively small training set size (100 samples).
By increasing $\fmax$, the FOM state gets less prone to be represented by a compact encoding; still, as the number of latent states increases, the accuracy is enhanced.

\subsection{Test Case 2: Unsteady Navier-Stokes}

\begin{figure}
    \includegraphics[width=\textwidth]{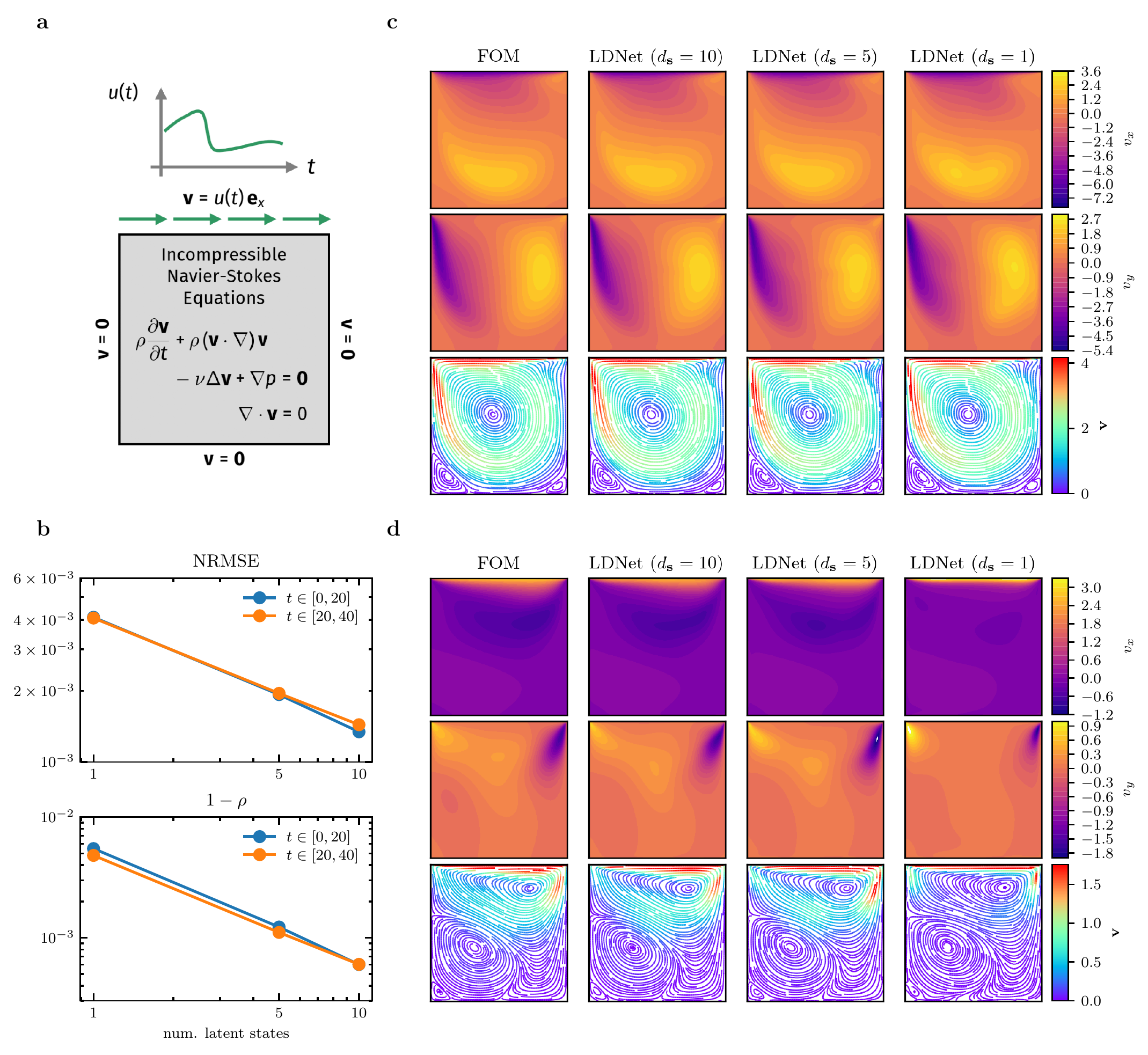}
    \caption{
        \textbf{Test Case 2. }
        \figpt{a}: Computational domain and equations of the FOM.
        \figpt{b}: Error metrics (NRMSE and Pearson dissimilarity) of LDNets for different number of latent variables (1, 5 and 10). 
        The training dataset consists of 80 simulations with $\timeMax = 20$, while the test dataset comprises 200 simulations with $\timeMax = 40$.
        The blue lines refer to the testing error obtained in the interval $t \in [0, 20]$ (that is the same interval seen during training), while orange lines refer to the testing error in the interval $t \in [0, 40]$.
        \figpt{c}: A snapshot of the velocity field within the interval $t \in [0, 20]$ (interpolation interval) of a testing sample.
        \figpt{d}: A snapshot of the velocity field within the interval $t \in [20, 40]$ (extrapolation interval) of a testing sample.
        }
    \label{fig:NS_grid}
\end{figure}

The 2D lid-driven cavity is a well-known benchmark problem in fluid dynamics \cite{zienkiewicz2005finite}, which may exhibit a wide range of flow patterns and vortex structures when increasing the Reynolds number.
We challenge LDNets in learning an unsteady version of the lid-driven cavity problem, where the velocity prescribed on the top portion of the boundary ($\GammaTop$) is a time-dependent input $u(\timevar)$ (see Fig.~\ref{fig:NS_grid}a). 
In particular, as reported in Sec.~\ref{sec:SI:results:ADR:sampling}, we consider a Gaussian Process Distribution with mean $\mu = 0$, standard deviation $\sigma = 5$ and characteristic time-scale $\tau = 5$.
During the simulations, the Reynolds number varies over time by reaching peaks of nearly 1500.
This problem is challenging also because of discontinuities in the velocity field at the two top corners.
The goal here is to predict the velocity field (that is, we set $\outputField(\spacevar,\timevar) = \velocity(\spacevar,\timevar)$) for each prescribed $u(\timevar)$:
\begin{equation} \label{eqn:FOM_NS}
        \begin{aligned}
            &\rho  \frac{\partial \velocity}{\partial t} + \rho \left( \velocity \cdot \nabla \right)\velocity - \mu \Delta\velocity + \nabla p = \mathbf{0}
            \quad & & \mathbf{x} \in \Omega, \, t \in (0, T] ,\\
            &\nabla \cdot \velocity = 0
            \quad & & \mathbf{x} \in \Omega, \, t \in (0, T] ,\\
            &  \velocity = u(\timevar) \mathbf{e}_x
            \quad & & \mathbf{x} \in \GammaTop, \, t \in (0, T] ,\\
            &  \velocity = \mathbf{0}
            \quad & & \mathbf{x} \in \partial\Omega \setminus \GammaTop, \, t \in (0, T] ,\\
            & \velocity = \mathbf{0}
            \quad & & \mathbf{x} \in \Omega, t = 0 ,\\
        \end{aligned}
\end{equation}
where the dependence of the velocity $\velocity$ and pressure $p$ on space and time is understood.
As shown in \cite{regazzoni2022usmnets}, a simple quadratic loss function is not adequate for capturing small vortex structures, because of their small impact, compared to medium- and large-scale structures, to the loss function.
Therefore, we use the following goal-oriented metric, where we denote by $\velocity$ and $\hat{\velocity}$ the reference and predicted velocities, respectively:
\begin{equation} \label{eqn:metric_NS}
    \error(\velocity, \hat{\velocity}) = 
    \frac{\| \velocity - \hat{\velocity} \|^2}{\NSrefVel^2}
    + \gamma \left\| \frac{\velocity}{\epsilon + \|\velocity\|} - \frac{\hat{\velocity}}{\epsilon + \|\hat{\velocity}\|} \right\|^2 
\end{equation}
with hyperparameters $\gamma$ and $\epsilon \ll 1$, and where $\NSrefVel$ is a reference velocity magnitude.
Specifically, we pick $\gamma = \num{1e-1}$ and $\epsilon = \num{1e-4}$.
The second term of the metric \eqref{eqn:metric_NS} allows to match the flow direction, even in the regions of small flow magnitude.

We generate training data through a FEM-based solver of \eqref{eqn:FOM_NS}, on a $100 \times 100$ triangular grid, accounting for nearly 91K degrees of freedom. 
We employ Taylor-Hood elements in the FEM solver, i.e. $\mathbb{P}_2$ Finite Elements for $\velocity$ and $\mathbb{P}_1$ Finite Elements for $p$.
We consider a semi-implicit time discretization with $\Dt = \num{2e-1}$.
The training and validation datasets consist of 80 simulations with $T = 20$, where 200 points are uniformly sampled in $\Omega = (0, 1)^2$.
Regarding the testing data, we run 200 simulations with $T = 40$, a time span that is twice as long as the one of the training set, and we uniformly take 400 points in space from the domain $\Omega = (0, 1)^2$.
To train LDNets, we take 100 evenly distributed snapshots in time, and we randomly take 200 points in space for each time-step.

We train three LDNets, by increasing the number of latent states from 1 to 5 and 10.
Specifically, we fix the number of latent states and we perform hyperparameters tuning (see Sec.~\ref{sec:SI:methods:hypertuning}) by monitoring the following goal-oriented loss function \eqref{eqn:metric_NS}.
We define in Tab.~\ref{tab:NS:hyperparameters} the initial ranges and final values of the hyperparameters for different numbers of latent states (1, 5 and 10).

\begin{sidewaystable}
    \centering
    \begin{tabular}{ll|cccccc|cc}
        \toprule
        \multicolumn{2}{l|}{\textbf{Test cases}}  &\multicolumn{6}{c|}{\textbf{Hyperparameters}}                                                                        & \multicolumn{2}{c}{\textbf{Trainable parameters}} \\
                &                        &\multicolumn{2}{c}{$\ROMrhs$}       & \multicolumn{2}{c}{$\ROMobs$}    & $\DtREF$            & $\weightREGrhs$, $\weightREGobs$  & $\ROMrhs$ & $\ROMobs$ \\
                &                        & layers          & neurons          & layers         & neurons         &                     &                                   & \# param. & \# param. \\ \midrule \multicolumn{2}{l}{\textbf{Test Case $\NUMstateROM = 1$}} & \multicolumn{8}{l}{}\\
        \multicolumn{2}{l|}{tuning      }& \myrange{1}{6}  & \myrange{5}{35}  & \myrange{1}{6} & \myrange{5}{35} & $[10^{-1}, 10^{1}]$ & 0                                 &           &          \\
        \multicolumn{2}{l|}{final       }& 2               & 7                & 4              & 24              & 5.4                 & 0                                 & 85        & 1'970      \vspace{.25cm} \\ \midrule \multicolumn{2}{l}{\textbf{Test Case $\NUMstateROM = 5$}} & \multicolumn{8}{l}{}\\
        \multicolumn{2}{l|}{tuning      }& \myrange{1}{6}  & \myrange{5}{35}  & \myrange{1}{6} & \myrange{5}{35} & $[10^{-1}, 10^{1}]$ & 0                                 &           &          \\
        \multicolumn{2}{l|}{final       }& 2               & 27               & 4              & 33              & 8.6                 & 0                                 & 1'085     &  3'731      \vspace{.25cm} \\ \midrule \multicolumn{2}{l}{\textbf{Test Case $\NUMstateROM = 10$}} & \multicolumn{8}{l}{}\\
        \multicolumn{2}{l|}{tuning      }& \myrange{1}{6}  & \myrange{5}{35}  & \myrange{1}{6} & \myrange{5}{35} & $[10^{-1}, 10^{1}]$ & 0                                 &           &          \\
        \multicolumn{2}{l|}{final       }& 3               & 19               & 5              & 28              & 9.2                 & 0                                 & 1'188     &  3'698      \vspace{.25cm} \\
        \bottomrule
    \end{tabular}
    \caption{Test Case 2: hyperparameters ranges and selected values. See text for details.}
    \label{tab:NS:hyperparameters}
\end{sidewaystable}

We report in Tab.~\ref{tab:NS:results} the NRMSE and Pearson dissimilarity values on the testing set for $\NUMstateROM=1$, $\NUMstateROM=5$ and $\NUMstateROM=10$.
We notice that the size of both $\ROMrhs$ and $\ROMobs$ architectures increases while the testing error of the LDNets decreases with respect to the number of latent states.
Similarly to Test Case 1, as the number of latent states increases the LDNets are more and more efficient in discovering an effective compact representation of the system state and thus providing reliable predictions (Figs.~\ref{fig:NS_grid}b and \ref{fig:NS_grid}c).
Still, small NRMSEs are attained even with a small number of latent states.

Furthermore, we challenge the trained LDNets in predicting the flow evolution even on a longer time horizon than that considered in the training dataset (specifically, twice as long), that is for $t \in (20, 40]$, when the training samples are generated for $t \in [0, 20]$.
Remarkably, we observe a negligible propagation of the approximation error along the prolonged time frame, making the trained LDNets reliable also for time-extrapolation (Figs.~\ref{fig:NS_grid}b and \ref{fig:NS_grid}d, Tab.~\ref{tab:NS:results}).
This is a remarkable achievement, considering that the dynamical system at hand does not present a periodic or quasi-periodic regime.

\begin{table}
    \centering
    \begin{tabular}{l|rrr}
        \toprule
        Number of latent states & 1 & 5 & 10 \\
        \midrule
        NRMSE\textsubscript{test} ($0 < t < 40$)   & \num{4.08e-03} & \num{1.94e-03} & \num{1.39e-03} \\
        NRMSE\textsubscript{test} ($0 < t < 20$)   & \num{4.10e-03} & \num{1.93e-03} & \num{1.34e-03} \\
        NRMSE\textsubscript{test} ($20 < t < 40$)  & \num{4.07e-03} & \num{1.95e-03} & \num{1.44e-03} \\
        $1 - \rho_{\mathrm{test}}$ ($0 < t < 40$)  & \num{5.14e-03} & \num{1.16e-03} & \num{6.01e-04} \\
        $1 - \rho_{\mathrm{test}}$ ($0 < t < 20$)  & \num{5.48e-03} & \num{1.23e-03} & \num{6.00e-04} \\
        $1 - \rho_{\mathrm{test}}$ ($20 < t < 40$) & \num{4.81e-03} & \num{1.11e-03} & \num{6.02e-04} \\
        \bottomrule
    \end{tabular}
    \caption{Test Case 2: test accuracy metrics for LDNets trained with an increasing number of latent states (1, 5 and 10).}
    \label{tab:NS:results}
\end{table}

\begin{figure}
    \includegraphics[width=\textwidth]{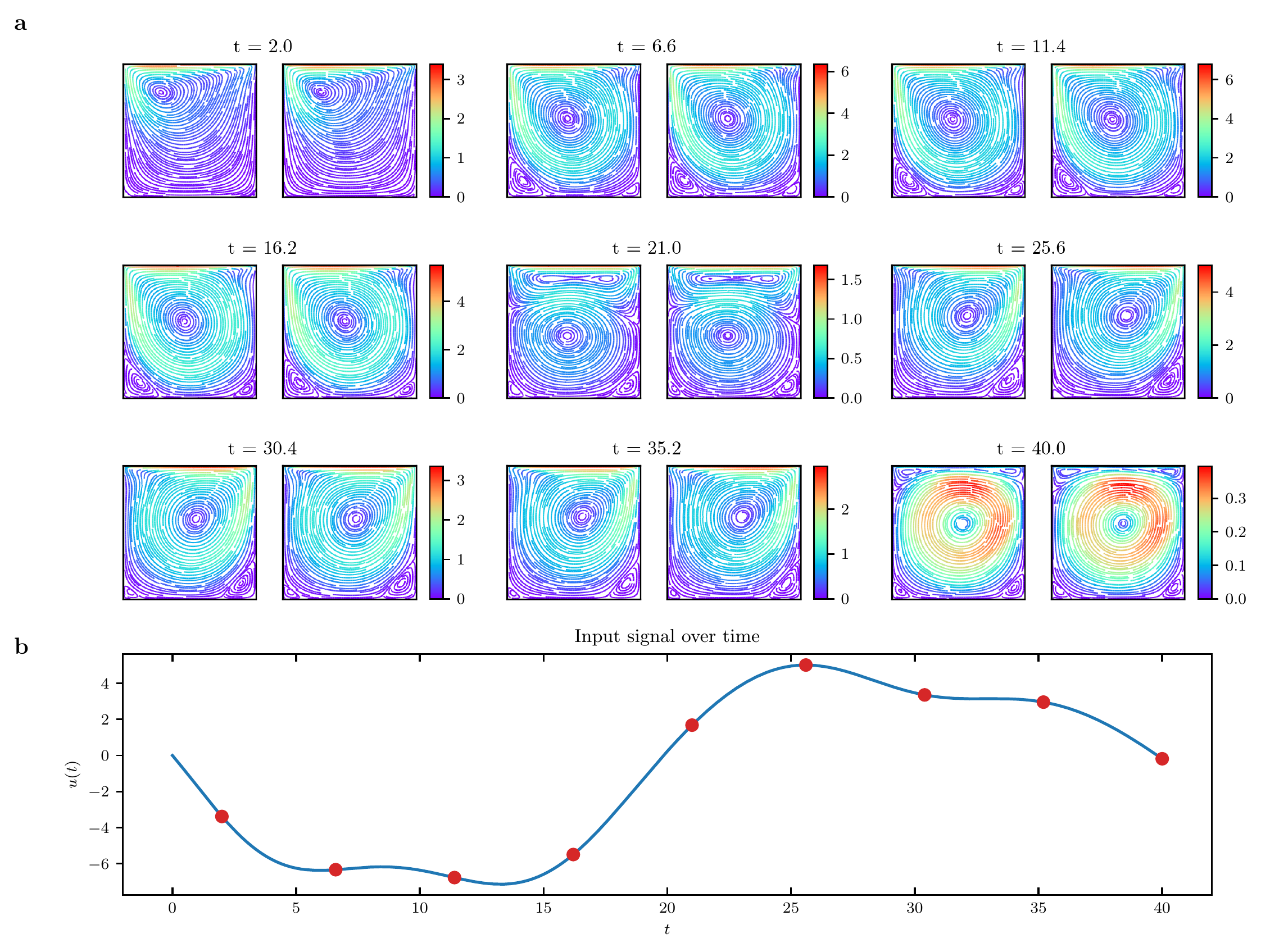}
    \caption{Test Case 2.
             \figpt{a}: Streamlines of the velocity field $\velocity$ for one testing sample over time ($t \in [0, 40]$). Horizontal and vertical axis refer to space in the domain $\Omega = (0, 1)^2$. For each subplot, the left one represents the FOM while the right one represents the LDNet approximation for $\NUMstateROM = 10$.
             \figpt{b}: Input signal $u(\timevar)$ over time applied to the top portion of the boundary $\GammaTop$.}
    \label{fig:NS_figures_samples}
\end{figure}

We depict in Fig.~\ref{fig:NS_figures_samples} the time evolution of the streamlines associated with a specific sample belonging to the testing set, along with the corresponding input signal $u(\timevar)$.
We see that the approximation provided by LDNets with 10 latent variables has an excellent agreement with the FOM. 

\subsection{Test Case 3: Aliev-Panfilov electrophysiology model}

We consider the Aliev-Panfilov (AP) model \cite{aliev1996simple}, a nonlinear system of PDEs describing the propagation of the electrical potential $\APstate(x,t)$ in an excitable tissue.
The AP model envisages a recovery variable $\APrecovery(x,t)$ that tracks the refractoriness of the tissue by modulating the repolarization phase.
The model, supplemented with homogeneous Neumann boundary conditions (encoding electrical insulation) and zero initial conditions for both the variables, reads
\begin{equation} \label{eqn:FOM_AP}
        \begin{aligned}
            &\frac{\partial \APstate}{\partial t} 
            - \APD \frac{\partial^2 \APstate}{\partial x^2} 
            =
            \APK \APstate (1 - \APstate)(\APstate - \APalpha)
            - \APstate \APrecovery 
            + \APIapp
            \quad & & x \in (0, \APL), \, t \in (0, T] ,\\
            &\frac{\partial \APrecovery}{\partial t} 
            =
            \left(\APgamma + \frac{\APmuOne \APrecovery}{\APmuTwo + \APstate}\right)
            \left(-\APrecovery - \APK \APstate (\APstate - \APb - 1)\right)
            \quad & & x \in (0, \APL), \, t \in (0, T] ,\\
            & \frac{\partial \APstate(   0,t)}{\partial x} 
            = \frac{\partial \APstate(\APL,t)}{\partial x} 
            = 0
            \quad & & t \in (0, T] ,\\
            & \APstate(x, 0) = \APrecovery(x, 0) = 0,
            \quad & & x \in (0, \APL) .\\
        \end{aligned}
\end{equation}
with parameters
$\APD = \SI{0.1}{\milli\meter\per\milli\second}$, 
$\APK = 8$, 
$\APalpha = 0.1$, 
$\APgamma = 0.02$, 
$\APmuOne = 0.2$, 
$\APmuTwo = 0.3$, 
$\APb = 0.15$, 
$\APL = \SI{100}{\milli\meter}$, 
$\timeMax = \SI{500}{\milli\second}$. Note that $z$ is a non-dimensional potential (which can be mapped to its physiological values by the relationship $(100 z - {80})\si{\milli\volt}$) and the model is rescaled with respect to the time constant $\tau=\SI{12.9}{\milli\second}$ (for further details, see \cite{goktepe2009computational}).

The excitation-propagation process is triggered by an external stimulus $\APIapp(x,t)$, applied at two stimulation points, respectively located at $x = 1/4 \APL$ and $x = 3/4 \APL$, and consisting of square impulses, to mimic the action of a (natural of artificial) pacemaker.
The input of the model is given by $\inputSign(t) = (\APIapp(\APxapp{1},t), \APIapp(\APxapp{2},t))$.
To generate the training samples, we randomly trigger the applied stimuli, either in correspondence of $\APxapp{1}$, $\APxapp{2}$ or both points, by randomly picking the stimulation times.
For the approximation of the AP model, we employ the finite difference method both in space and time, on a regular grid with 800 points in space and $10^5$ time steps. 
Then, we subsample the space-time grid by retaining 100 points in space and 500 time instants.

The AP model solution features the fast-slow dynamics of a cardiac action potential (steep depolarization fronts followed by slow repolarization of the electrical potential to its resting value) and the wavefront propagation in space generating collisions of waves from different stimulation points.
These features make this problem a challenging test case for comparing the proposed method against popular approaches to learning space-time dynamics of complex systems.

We compare LDNets with state-of-the-art approaches in which dimensionality reduction is achieved by training an auto-encoder (AE) on a discrete representation of the output $\APstate(\cdot,t)$.
Once trained, the encoder is employed to compute the trajectories of the latent states throughout the training set, and the dynamics in the latent space is learned either through an ODE-Net \cite{chen2018neural} or an LSTM \cite{hochreiter1997long}.
We denote the resulting models by AE/ODE and AE/LSTM, respectively.
Then, we further train the NN that tracks the dynamics of the latent states simultaneously with the decoder, that is in an end-to-end (e2e) fashion, and we denote the resulting models by AE/ODE-e2e and AE/LSTM-e2e, respectively.
Furthermore, we benchmark LDNets against a classical method of model-order reduction of PDE models, namely the POD-DEIM method \cite{pagani2018numerical,DalSantoManzoniPaganiQuarteroni}.
These methods are described in detail in Appendix~\ref{app:alternative_methods}.

We challenge LDNets and the above-mentioned methods in the task of predicting the space-time dynamics of the target value $\outputField(x,\timevar) = \APstate(x,\timevar)$, given the time series of impulses in the two stimulation points.
In order to ensure a fair comparison, for all the methods that require a choice of hyperparameters we use an algorithm for their automatic tuning, as described in Sec.~\ref{sec:SI:methods:hypertuning}.
The ranges used for tuning and the final hyperparameter values are reported in Tab.~\ref{tab:AP:AE} and \ref{tab:AP:LDNet}.
In this test case, we employ the technique described in Sec.~\ref{sec:methods:equilibrium} to impose in a strong manner the equilibrium condition of the initial state when $\inputSign(t) = \mathbf{0}$.
To achieve a significant dimensionality reduction, we set a maximum number of 12 latent variables both for auto-encoder-based methods and for LDNets.
We notice that the hyperparameter tuning algorithm selects the maximum number of latent states (i.e. $\NUMstateROM = 12$) for all the methods.
For the POD-DEIM method, we test different number $\NUMstateROM$ of POD modes (i.e., basis functions) for the state and the nonlinear term, ranging from 12 to 60.

\begin{table}
	\centering
	\begin{tabular}{l|cccccccc}
		\toprule
               &\multicolumn{8}{c}{\textbf{Hyperparameters}} \\
               & $\NUMstateROM$   & \multicolumn{2}{c}{$\ROMenc$, $\ROMdec$} & \multicolumn{2}{c}{$\ROMrhs$}          & $\DtREF$            & $\weightREGenc$, $\weightREGdec$ & $\weightREGrhs$      \\ 
		       &                  & layers           & neurons               & layers           & neurons             &                     &                                  &                      \\ \midrule \multicolumn{4}{l}{\textbf{AE/ODE}}\\
        tuning & \myrange{4}{12}  &  \myrange{1}{4}  & \myrange{4}{80}       & \myrange{1}{10}  & \myrange{4}{40}     & $[10^{1}, 10^{3}]$  & $[10^{-5}, 10^{-2}]$             & $[10^{-5}, 10^{-1}]$ \\
        final  & 12               &  1               & 75                    & 4                & 38                  & $\num{1.80e+01}$    & $\num{6.89e-03}$                 & $\num{2.80e-02}$     \vspace{.25cm} \\ \midrule \multicolumn{4}{l}{\textbf{AE/LSTM}}\\
        tuning & \myrange{4}{12}  &  \myrange{1}{4}  & \myrange{4}{80}       &                  &                     &                     & $[10^{-5}, 10^{-2}]$             & $[10^{-5}, 10^{0}]$  \\
        final  & 12               &  1               & 75                    &                  &                     &                     & $\num{6.89e-03}$                 & $\num{4.81e-01}$     \\
    \bottomrule
	\end{tabular}
	\caption{Test Case 3: hyperparameters ranges and selected values for the AE/ODE and AE/LSTM models. 
    For the encoder (respectively, the decoder) the number of neurons refers to the first (respectively, last) hidden layer.
    In the other layers, the number of neurons is linearly varied to connect the first (respectively, last) hidden layer to the layer of latent variables.
    }
	\label{tab:AP:AE}
\end{table}

\begin{table}
	\centering
    \begin{tabular}{l|ccccccc}
		\toprule
               &\multicolumn{7}{c}{\textbf{Hyperparameters}} \\
               & $\NUMstateROM$   &\multicolumn{2}{c}{$\ROMrhs$}     & \multicolumn{2}{c}{$\ROMobs$}       & $\DtREF$            & $\weightREGrhs$, $\weightREGobs$  \\
		       &                  & layers         & neurons         & layers          & neurons           &                     &                                   \\ \midrule \multicolumn{4}{l}{\textbf{LDNet}} \\
        tuning & \myrange{4}{12}  & \myrange{1}{3} & \myrange{4}{15} & \myrange{1}{5}  & \myrange{4}{20}   & $[10^{1}, 10^{3}]$  & $[10^{-5}, 10^{-2}]$              \\
        final  & 12               & 1              & 5               & 5               & 17                & $\num{2.05e+02}$    & $\num{4.70e-03}$                  \\
        \bottomrule
	\end{tabular}
	\caption{Test Case 3: hyperparameters ranges and selected values for the LDNet.}
	\label{tab:AP:LDNet}
\end{table}

\begin{figure}
    \includegraphics[width=\textwidth]{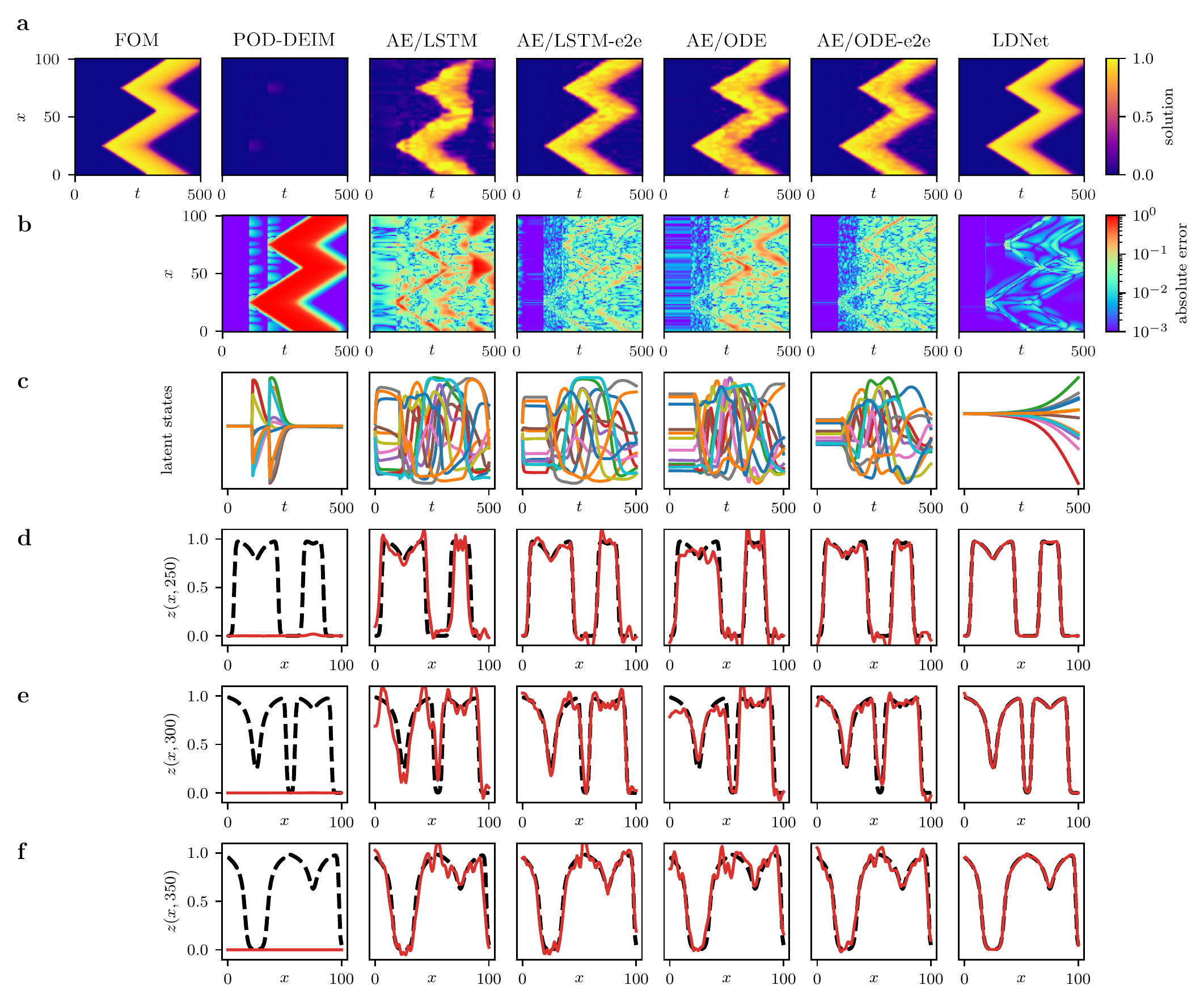}
    \caption{
        \textbf{Test Case 3: Methods comparison.}
        We compare the results obtained with different methods for a sample belonging to the test dataset.
        The left-most column reports the FOM solution of the AP model (the abscissa denotes time, the ordinate denotes space).
        For each method we report:
        \figpt{a} the space-time solution;
        \figpt{b} the space-time error with respect to the FOM solution;
        \figpt{c} the time-evolution of the 12 latent variables;
        \figpt{d}-\figpt{e}-\figpt{f} three snapshots of the space-dependent output field at $t = 250$, $300$ and $350$, in which we compare the predicted solution (red solid line) with the FOM solution (black dashed line). 
        }
    \label{fig:AP_graph_comparison}
\end{figure}

\begin{figure}
    \includegraphics[width=\textwidth]{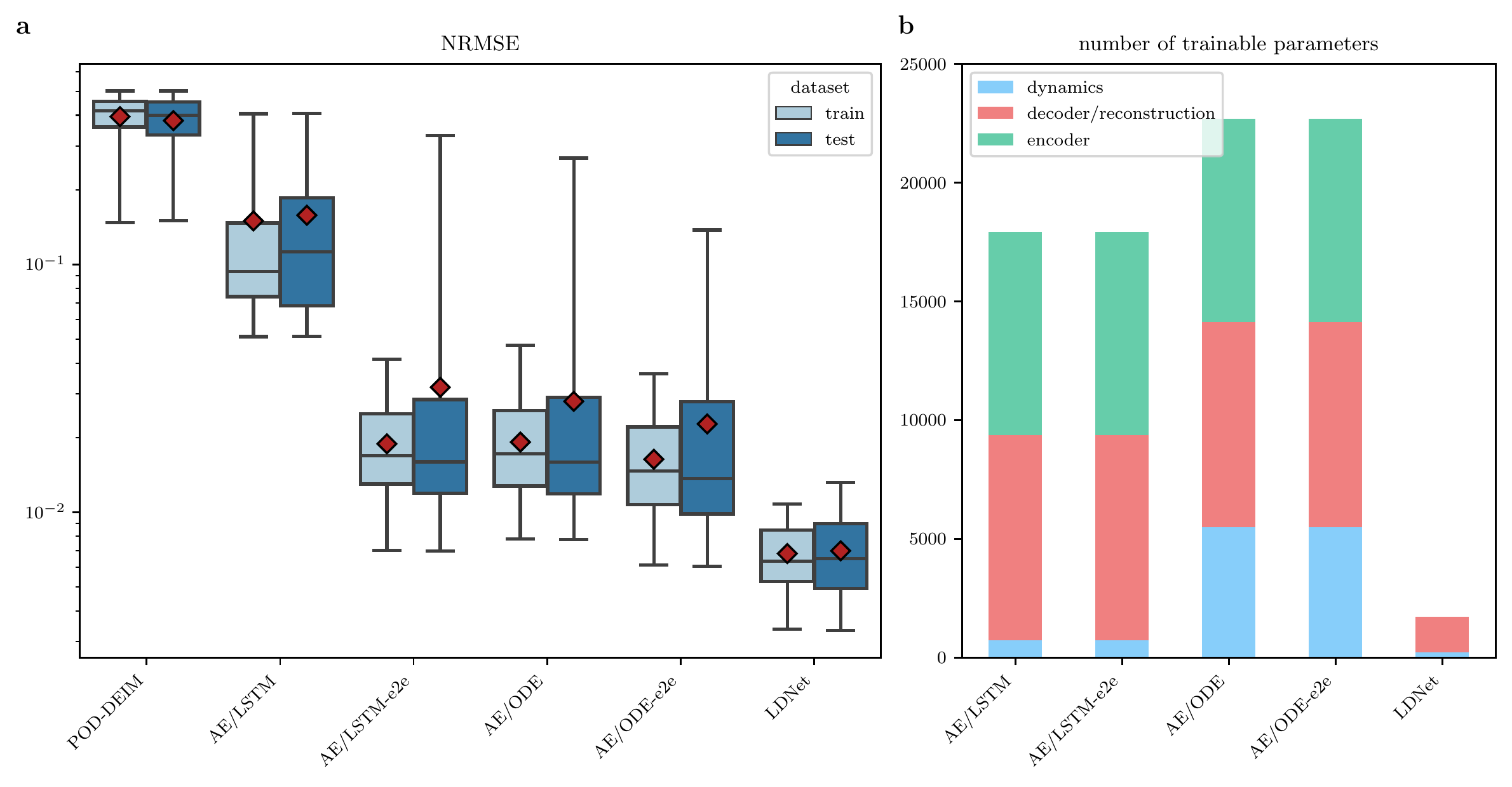}
    \caption{
        \textbf{Results of Test Case 3.}
        \figpt{a}: Boxplots of the distribution of the testing (blue) and training (light blue) errors obtained with each method. 
        The red diamonds represent the average error on each dataset.
        \figpt{b}: Number of trainable parameters of each method.
        The bin \textit{encoder} is present only for auto-encoder-based methods, but not for LDNets.
        The bin \textit{dynamics} refers to the NN that evolves the latent states.
        The POD-DEIM method is not included, as it does not envisage a training stage.
        }
    \label{fig:AP_errors}
\end{figure}

In Tab.~\ref{tab:AP_errors} we report training and testing errors obtained for the different methods, along with the number of trainable parameters.
The results of this comparison are also reported in Figs.~\ref{fig:AP_graph_comparison}--\ref{fig:AP_errors}.

\begin{table}
	\centering
    \begin{tabular}{l|r|r|r|r|r|r}
		\toprule
                                       & \multicolumn{2}{c|}{NRMSE}           &  \multicolumn{4}{c}{Number of trainable parameters}   \\
                                       & training         &  testing          & $\ROMenc$ & $\ROMdec$, $\ROMobs$ & $\ROMdyn$ & total  \\ \midrule
        POD-DEIM ($\NUMstateROM = 12$) &  \num{4.05e-01}  &  \num{3.92e-01}   &           &                      &           &        \\
        POD-DEIM ($\NUMstateROM = 24$) &  \num{3.59e-01}  &  \num{3.47e-01}   &           &                      &           &        \\
        POD-DEIM ($\NUMstateROM = 36$) &  \num{1.71e-01}  &  \num{1.62e-01}   &           &                      &           &        \\
        POD-DEIM ($\NUMstateROM = 48$) &  \num{7.48e-02}  &  \num{7.57e-02}   &           &                      &           &        \\
        POD-DEIM ($\NUMstateROM = 60$) &  \num{2.97e-02}  &  \num{2.90e-02}   &           &                      &           &        \\
             AE/LSTM                   &  \num{1.90e-01}  &  \num{1.98e-01}   &    8'562  &   8'651              &    720    & 17'933 \\
         AE/LSTM-e2e                   &  \num{2.05e-02}  &  \num{5.87e-02}   &    8'562  &   8'651              &    720    & 17'933 \\
              AE/ODE                   &  \num{2.09e-02}  &  \num{4.58e-02}   &    8'562  &   8'651              &  5'484    & 22'697 \\
          AE/ODE-e2e                   &  \num{1.78e-02}  &  \num{3.37e-02}   &    8'562  &   8'651              &  5'484    & 22'697 \\
               LDNet                   &  \num{7.09e-03}  &  \num{7.37e-03}   &        0  &   1'480              &    228    &  1'708 \\
        \bottomrule
	\end{tabular}
	\caption{Test Case 3: Training and testing errors obtained with the different methods and number of trainable parameters.}
	\label{tab:AP_errors}
\end{table}

\begin{figure}
    \includegraphics[width=\textwidth]{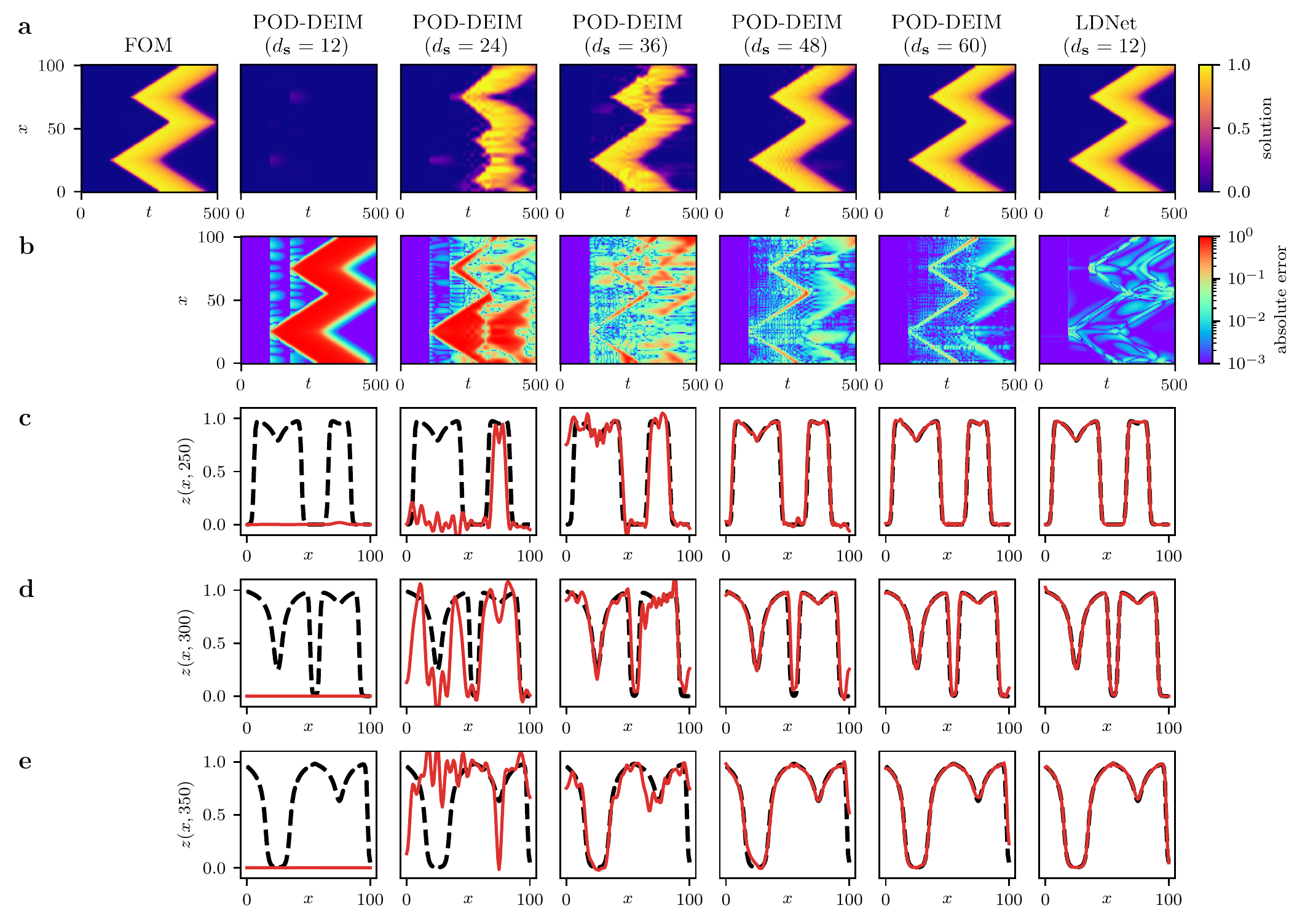}
    \caption{
        We compare the results obtained with the POD-DEIM method, for an increasing number of considered modes (reported in the titles), against the results obtained with our proposed method.
        The figure shows the predictions obtained for a sample belonging to the test dataset.
        The left-most column reports the FOM solution of the AP model (the abscissa denotes time, the ordinate denotes space).
        \figpt{a} the space-time solution;
        \figpt{b} the space-time error with respect to the FOM solution;
        \figpt{c}-\figpt{d}-\figpt{e} three snapshots of the space-dependent output field at $t = 250$, $300$ and $350$, in which we compare the predicted solution (red solid line) with the FOM solution (black dashed line). 
        }
    \label{fig:AP_graph_comparison_RB}
\end{figure}

\begin{figure}
    \centering
    \includegraphics[width=0.6\textwidth]{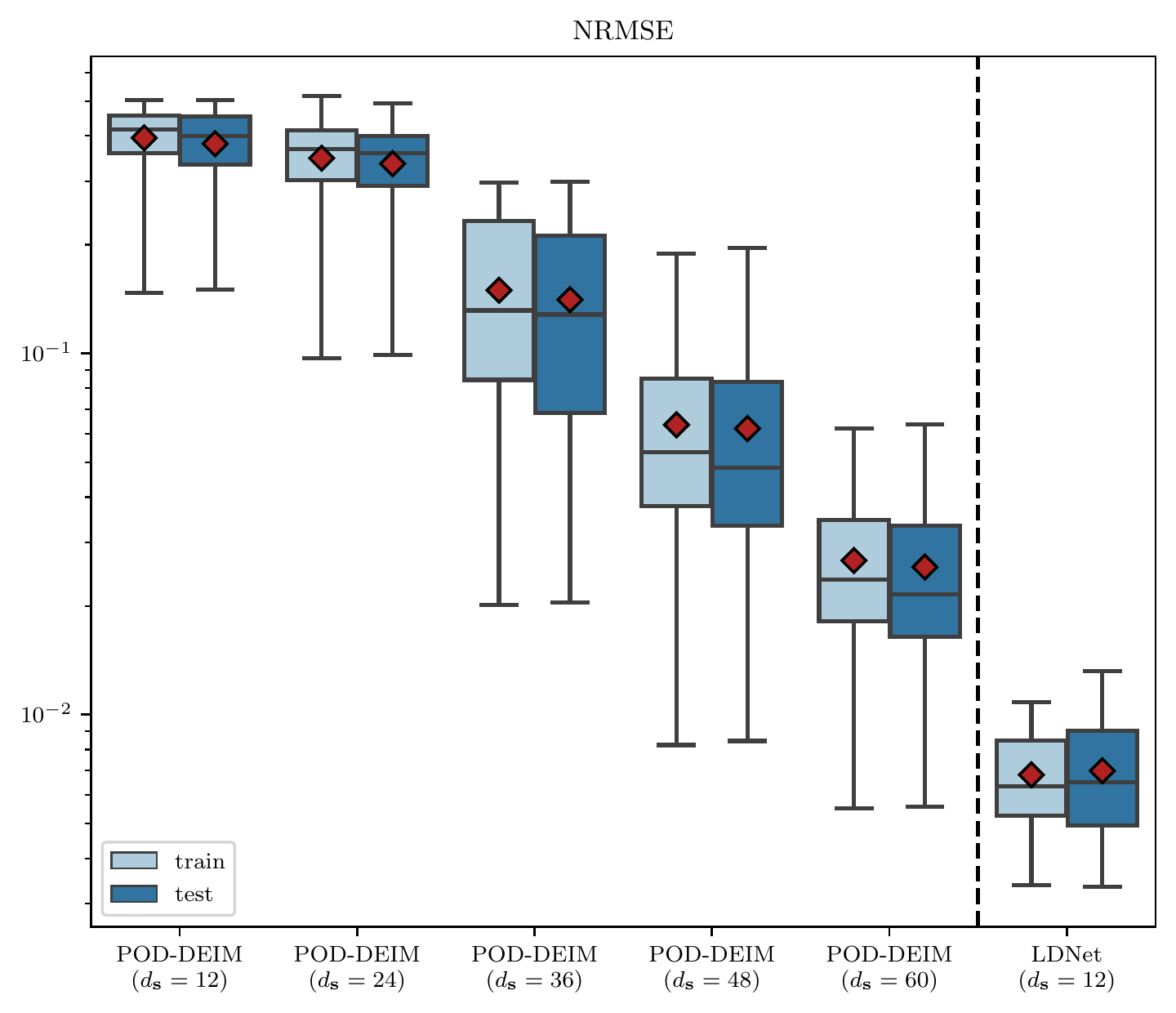}
    \caption{
        We compare the results obtained with the POD-DEIM method, for an increasing number of considered modes, against the results obtained with our proposed method.
        Specifically, we report boxplots of the distribution of the testing (blue) and training (light blue) errors obtained with each method. 
        The red diamonds represent the average error on each dataset.
        }
    \label{fig:AP_errors_RB}
\end{figure}

We observe that the results obtained with the POD-DEIM method using 12 modes are unsatisfactory, compared to the other methods, confirming the importance of adopting nonlinear dimensional reduction techniques for this class of traveling-front problems.
Indeed, due to the presence of traveling fronts, this problem features a slow decay of the Kolmogorov $n$-width \cite{quarteroni2015reduced}, that reflects in a poor accuracy of the electrical potential reconstruction given by the POD-DEIM method when 12 modes are used.
Then, we consider higher number of modes, namely 24, 36, 48 and 60.
We report in Fig.~\ref{fig:AP_graph_comparison_RB}, Fig.~\ref{fig:AP_errors_RB} and Tab.~\ref{tab:AP_errors} the results.
By increasing the number of considered modes, the accuracy increases.
With 60 modes, e.g., the POD-DEIM method is more accurate than auto-encoder-based methods with 12 latent variables, while LDNets with 12 latent variables are still more accurate.
Nonetheless, we observe that an increase of number of modes in the POD-DEIM method is accompanied by a significant increase in the computational cost, since the computational complexity scales with the cube of $\NUMstateROM$. 
In addition, the POD-DEIM model is solved on the same temporal discretization as the high-fidelity one, which represents a strong limitation with respect to the other methods described in this paper. 

A better accuracy is achieved by both auto-encoder-based methods and by LDNets, thanks to their ability to express a nonlinear relationship between the latent states and the solution.
Still, LDNet outperforms the other methods, with a testing NRMSE equal to $\num{7e-3}$ and with a much lower overfitting.
The testing NRMSE of auto-encoder-based methods is nearly 5 times larger than with LDNets or more.
Remarkably, our method achieves better accuracy with significantly fewer trainable parameters: auto-encoder-based methods require more than tenfold the number of parameters.
This testifies to the good architectural design of LDNets.
	\section*{Discussion} \label{sec:discussion}

We have presented LDNets, a novel class of NNs that learn in a data-driven manner the evolution of systems exhibiting spatio-temporal dynamics in response to external inputs.
An LDNet is trained in a supervised way from observations of input-output pairs, which can either come from experimental measurements or be synthetically generated through the numerical approximation of mathematical models of which one seeks a surrogate or reduced-order model.
This latter case is the one considered in this manuscript to demonstrate the capabilities of the proposed method.

LDNets provide a paradigm-shift from state-of-the-art methods based on dimensionality reduction (e.g., exploiting POD or auto-encoders) of a high-dimensional discretization of the system state. 
Specifically, LDNets automatically discover a compact representation of the system state, without necessitating the explicit construction of an encoder.
This enables the training algorithm to select a compact representation of the state that is functional not only in reconstructing the space-dependent field for each time instant, but also in predicting its dynamics; an auto-encoder, conversely, when trained, extracts features on a purely statistical basis, being agnostic of the importance of each feature in determining the evolution of the system.

Unlike standard approaches that reconstruct a high-dimensional discretization of the output, corresponding e.g. to evaluations at the vertices of a computational mesh, our approach is in this sense meshless.
The reconstruction NN is indeed queried for each point in space independently. 
This design principle gives LDNets several benefits.
First, the meshless nature of LDNets combined with the automatic discovery of the latent space allows them to operate in a low-dimensional space without ever going through a high-dimensional discretization, as auto-encoder-based methods do.
This makes LDNets very lightweight structures, easy to train, and not prone to overfitting.
The LDNet architecture enables the sharing of the trainable parameters needed to evaluate the solution at different points (that is, the same weights are employed regardless of the query point).
The low overfitting of LDNets is thus not surprising, as weight-sharing is often the key of good generalization properties of many architectures, such as CNNs and RNNs \cite{goodfellow2016deep}.
Second, it provides a continuous representation of the output, and, thus, allows for additional and possibly physics-informed terms to be introduced into the loss function \cite{raissi2019physics}, opening up countless possibilities for extending the purely black-box method proposed in this paper to grey-box approaches.
Third, the loss function can be defined by stochastically varying the points in space at which the error is evaluated (see Test Case 2), thus lightening the computational burden associated with training.
Note that this is not possible when the model returns the entire batch of observations.
This aspect also opens up to multiple developments, such as stochastic, minibatch-based training algorithms, or even adaptive refinements of the evaluation points, by sampling more densely where the error is larger.

The time-dynamics of LDNets is based on a recurrent architecture that is consistent, by construction, with the arrow of time.
This differentiates LDNets from other approaches in which time is seen as a parameter \cite{fresca2021comprehensive}, or approaches, based e.g. on DeepONets, that take as input the entire time-history of $\inputSign(\timevar)$ with a fixed length \cite{lu2021learning,oommen2022learning}.
The latter approaches do not easily allow for predictions over time frames longer than those used during training, or allow for time-extrapolation only in periodic o quasi-periodic problems \cite{zhu2022reliable}.
LDNets, on the other hand, allow predictions for arbitrarily long times.
We remark that the reliability of time-extrapolation is constrained by the characteristics of the problem at hand and the available training data.
For example, if the system is characterized by a divergent behavior such that, as time progresses, the state enters regions increasingly distant from the initial condition, then the reliability of the predictions is not guaranteed in time-extrapolation regimes.
When the system state remains bounded, however, the predictions of LDNets are significantly accurate even in time-extrapolation regimes, as showcased in Test Case 2.

LDNets represent, as proved by the results of this work, an innovative tool capable of learning spatio-temporal dynamics with great accuracy and by using a remarkably small number of trainable parameters.
They are able to discover, simultaneously with the system dynamics, compact representations of the system state, as shown in Test Case 1 where the Fourier transform of a sinusoidal signal is automatically discovered.
Once trained, LDNets provide predictions for unseen inputs with negligible computational effort (order of milliseconds for the considered Test Cases).
LDNets provide a new flexible and powerful tool for data-driven emulators that is open to a wide range of variations in the definition of the loss function (like, e.g., including physics-informed terms), in the training strategies, and, finally, in the NN architectures.
The comparison with state-of-the-art methods on a challenging problem, such as predicting the excitation-propagation pattern of a biological tissue in response to external stimuli, highlights the full potential of LDNets, which outperform the accuracy of existing methods while still using a significantly lighter architecture.

	\section*{Acknowledgements}

FR, SP and LD are members of the INdAM research group GNCS.
This project has been partially supported by the INdAM GNCS Project CUP\textunderscore E55F22000270001.
LD acknowledges the support of the FAIR (Future Artificial Intelligence Research) project, funded by the NextGenerationEU program (Italy) within the PNRR-PE-AI scheme (M4C2, investment 1.3, line on Artificial Intelligence).

	\begin{appendices}
		\section{Alternative methods} \label{app:alternative_methods}

In this appendix, we describe alternative methods to LDNets, whose performance is compared in Sec.~\ref{sec:PP:methods}.
These methods are based on a space discretization of the solution field.

We consider hence a space-discretization operator $\discrOperatorCD\colon\SEToutputField\to\mathbb{R}^{\NUMstateDISCR}$, with $\NUMstateDISCR \gg 1$.
The operator $\discrOperatorCD$ typically consists of a point-wise evaluation of $\outputField$ on a grid of points (e.g. the nodes of a computational mesh), or of an expansion with respect to a Finite Element basis or to a Fourier basis.
The subscript $h$ refers to the characteristic size associated with the discretization (e.g., the mesh element size for a mesh-based discretization; the sampling period for a discretization based on the discrete Fourier transform).
Hence, we write $\DISCRoutputField(\timevar) = \discrOperatorCD(\outputField(\cdot, \timevar)) \in \mathbb{R}^{\NUMstateDISCR}$.
The space-discretization operator is typically accompanied by a discrete-to-continuous operator $\discrOperatorDC\colon\mathbb{R}^{\NUMstateDISCR}\to\SEToutputField$, such that $\outputField(\cdot, \timevar) \simeq \discrOperatorDC(\DISCRoutputField(\timevar))$.

Let us consider, for simplicity, the case when the discretization operator is associated with the evaluation of the output field on a grid of points $\OBSspacevar_1, \OBSspacevar_2, \dots, \OBSspacevar_{\NUMobsPTS}$. 
Then we have 
\begin{equation*}
    \DISCRoutputField_i(\OBStimevar) = 
    (
        \outputField_i(\OBSspacevar_1, \OBStimevar),
        \outputField_i(\OBSspacevar_1, \OBStimevar), \dots,
        \outputField_i(\OBSspacevar_{\NUMobsPTS}, \OBStimevar)
    )
    \quad \text{for } i \in \trainSamples, \OBStimevar \in \trainTIMESoutputField{i}
\end{equation*}
In this case, we have $\NUMstateDISCR = \NUMoutputField\NUMobsPTS$.
We are thus restricting ourselves to the case when the training dataset is such that
$\trainPOINTSoutputField{i}{\OBStimevar} \equiv \trainPOINTSoutputField{}{} := (\OBSspacevar_1, \OBSspacevar_2, \dots, \OBSspacevar_{\NUMobsPTS})$. 
If, instead, observation points vary between samples or between time steps, interpolation would be required.

\subsection{Projection-based reduced-order models}

Projection-based reduced-order models (ROMs) allow the efficient simulation of complex spatio-temporal systems like \eqref{eqn:FOM} for many different queries of the input $\inputSign(\timevar)$ \cite{prud2002reliable,benner2005dimension,antoulas2005approximation,benner2015survey,MOR2021,hesthaven2022reduced}. 
They leverage Galerkin or Petrov-Galerkin projection of the PDE system onto problem-specific linear manifolds to reduce the complexity of high-fidelity discretization, deliberately rich in degrees of freedom for the sake of accurately approximating the PDE solution.
Here, to ease the notation, we only consider the case in which the model output coincides with the FOM state ($\outputField \equiv \stateFOM$), that is the observation operator is the identity function.
The case when the operator differs from the identity is a straightforward generalization of the one considered here.

Projection-based ROMs are intrusive, as they require manipulation of the discretization of FOM \eqref{eqn:FOM}. 
Their construction first involves dimensionality reduction techniques applied to pre-computed samples of the discretized FOM state:
\begin{equation*}
    \DISCRstateFOM(\timevar) = \discrOperatorCD(\stateFOM(\cdot, \timevar)) \in \mathbb{R}^{\NUMstateDISCR},
\end{equation*}
collected by numerically approximating the space-discrete counterpart of the FOM, for different samples of the input $\inputSign(\timevar)$ in the training set:
\begin{equation} \label{eqn:PFOM}
    \left\{ 
        \begin{aligned}
            &\frac{d}{dt} \DISCRstateFOM(\timevar) 
            = 
            \DISCRFOMrhs(
                \DISCRstateFOM(\timevar), 
                \inputSign(\timevar)
                )
            \qquad \text{for $ \timevar \in (0, \timeMax]$} \\
            &\DISCRstateFOM(\timevar) = \DISCRstateFOM_0 := \discrOperatorCD(\stateFOM_0),
        \end{aligned}
    \right.
\end{equation}
where $\DISCRFOMrhs$ is a suitable discretization of the differential operator $\FOMrhs$. 
Solution snapshots are stored into a matrix
\begin{equation*}
    Z = \{    \discrOperatorCD(\stateFOM_i(\cdot, \OBStimevar))  \}_{i \in \trainSamples}^{\OBStimevar \in \trainTIMESoutputField{i}}
    \in \mathbb{R}^{\NUMstateDISCR \times \NUMsnapshots}
\end{equation*}
from which $\NUMstateROM \ll \NUMstateDISCR$ basis functions $\boldsymbol{\phi}_k$, $k = 1,\ldots,\NUMstateROM$ are extracted by proper orthogonal decomposition (POD), or other linear dimensionality reduction techniques, like reduced-basis greedy algorithm \cite{quarteroni2015reduced,hesthaven2022reduced}. This allows expressing the discretized state compactly as
\begin{equation} \label{eqn:expansion}
    \DISCRstateFOM(\timevar) \approx V \statePROM(\timevar) = \sum_{k=1}^{\NUMstateROM} \statePROMcoeff(\timevar) \boldsymbol{\phi}_k,
\end{equation} 
where $V = [\boldsymbol{\phi}_1| \dots | \boldsymbol{\phi}_{\NUMstateROM}]$ is the transformation matrix, collecting the basis functions in its columns, and $\statePROM(\timevar) = [\statePROMcoeffIdx{1}(\timevar), \dots, \statePROMcoeffIdx{\NUMstateROM}(\timevar)]^T$ is the reduced state, a vector made of the coefficients $\statePROMcoeff$ associated with each basis function.
In this context, the solution manifold associated with the latent space is thus the column space of the matrix $V$.

Given a test manifold spanned by $\NUMstateROM$ basis functions $\{\boldsymbol{\psi}_k\}_{k=1}^{\NUMstateROM}$ (collected into a matrix $W = [\boldsymbol{\psi}_1| \dots | \boldsymbol{\psi}_{\NUMstateROM}]$), the projection of the dynamical system \eqref{eqn:PFOM} generates a system of $\NUMstateROM$ equations for the reduced state of the form:  
\begin{equation} \label{eqn:PROM}
    \left\{ 
        \begin{aligned}
            &\frac{d}{dt} \statePROM(\timevar) 
            = 
            W^T\DISCRFOMrhs(
                V \statePROM(\timevar), 
                \inputSign(\timevar)
                )
            \qquad \text{for $ \timevar \in (0, \timeMax]$}\\
            & \statePROM(0) = \statePROM_0 := V^T \DISCRstateFOM_0,
        \end{aligned}
    \right.
\end{equation}
In this formulation, computational efficiency hinges on the form of the discretized operator $\DISCRFOMrhs$. In the linear and affine cases, the dependence on the solution can be separated from the operator, generating a set or pre-computable terms that ensure computational efficiency in the resolution.  

In the case nonlinearities are present in the model, instead, $\DISCRFOMrhs$ requires the reduced state to be projected back onto the state space for its evaluation, before being projected onto the low-dimensional dynamics of \eqref{eqn:PROM}. This double projection makes the computational costs of \eqref{eqn:PROM} comparable to the ones of \eqref{eqn:PFOM}. Hyper-reduction techniques based on interpolation and linear dimensionality reduction techniques provide a computationally efficient alternative for evaluating nonlinearities \cite{quarteroni2015reduced,chaturantabut2010nonlinear,hesthaven2022reduced}. In our test case, we first separate the nonlinear part from the linear one: 
\begin{equation*}
\DISCRFOMrhs(\statePROM(\timevar),\inputSign(\timevar)) = \DISCRFOMrhs_{nl}(\statePROM(\timevar),\inputSign(\timevar)) + \DISCRFOMrhs_{l}(\statePROM(\timevar),\inputSign(\timevar)),
\end{equation*}
and we then rely on the discrete version of the empirical interpolation method (DEIM), which employs an interpolation-based projection of the nonlinearity onto the span of sparsely sampled basis functions, precomputed with POD \cite{chaturantabut2010nonlinear}. Specifically, the non-linear term $\DISCRFOMrhs_{nl}$ is approximated by
\begin{equation*}
\DISCRFOMrhs_{nl}(\statePROM(\timevar),\inputSign(\timevar)) \approx U (P^T U)^{-1} P^T \DISCRFOMrhs_{nl}(\statePROM(\timevar),\inputSign(\timevar)),
\end{equation*}
being $U$ the transformation matrix, collecting POD basis functions of the nonlinear term, and $P$ a sparse matrix that samples $\DISCRFOMrhs_{nl}$ on a subset of interpolation indices.
Here, we denote this model-order reduction method as POD-DEIM method.

The success and greatest limitation of projection-based ROM lies in being intrusively linked to high-fidelity discretization via a linear subspace. This, on the one hand, ensures the ROM consistency, with a reduced state directly mappable to the system state thanks to \eqref{eqn:expansion}, and convergence to the FOM as the dimensionality of the reduced manifold increases. On the other hand, if the solution of the parametric problem entails large variability, as in the case of advection-dominated problems, the accuracy-efficiency balance is compromised by the curse of dimensionality. 

\subsection{Methods based on auto-encoders}

For the reasons above, many researchers have shifted to the development of reduced-order models based on non-linear dimensionality reduction techniques, such as NN-based auto-encoders \cite{lee2020model,brunton2020machine,maulik2021reduced,fresca2021comprehensive,vlachas2022multiscale,oommen2022learning}. 
In this framework, to reduce the dimensionality of the discretized output field $\DISCRoutputField$, we train an auto-encoder, with latent code dimension $\NUMstateROM$, by considering the following minimization problem:
\begin{equation*}
    \begin{split}
        (\paramsROMenc^*, \paramsROMdec^*) = 
        \underset{\paramsROMenc, \paramsROMdec}{\operatorname{argmin}} 
        &\avsum_{i \in \trainSamples}
    \avsum_{\OBStimevar \in \trainTIMESoutputField{i}}
    \|\ROMdec(\ROMenc(\DISCRoutputField_i(\OBStimevar);\paramsROMenc);\paramsROMdec) - \DISCRoutputField_i(\OBStimevar) \|_{\DISCRoutputField}^2
    \\&+
    \weightREGenc \lossREG(\paramsROMenc)
    +
    \weightREGdec \lossREG(\paramsROMdec)
    \end{split}
\end{equation*}
where $\ROMenc$ and $\ROMdec$ (with trainable parameters $\paramsROMenc$ and $\paramsROMdec$, respectively) are the encoder and the decoder, respectively.
For normalization purposes, to evaluate distances in the discrete space we employ the rescaled euclidean norm $\| \cdot \|_{\DISCRoutputField}^2 := \| \cdot \|^2 / (\NUMobsPTS \, \NORMoutputField^2)$.

Once the auto-encoder is trained, i.e. the parameters $\paramsROMenc^*$ and $\paramsROMdec^*$ are available, we compute the latent codes associated with each training sample $i \in \trainSamples$ and each observation time $\OBStimevar \in \trainTIMESoutputField{i}$, that we denote as
\begin{equation*}
    \TARGETstateROM_i(\OBStimevar) = \ROMenc(\DISCRoutputField_i(\OBStimevar);\paramsROMenc^*) \in \mathbb{R}^{\NUMstateROM}.
\end{equation*}
At this stage, we train a second model (denoted by $\ROMdyn$, with trainable parameters $\paramsROMdyn$) to predict the dynamics of the latent codes $\TARGETstateROM_i(\timevar)$ as a function of the inputs $\VECinputSign$:
\begin{equation*}
    \begin{split}
        \paramsROMdyn^* = 
        \underset{\paramsROMdyn}{\operatorname{argmin}} 
        &\avsum_{i \in \trainSamples}
        \avsum_{\timevar \in \trainTIMESoutputField{i}}
        \|\ROMdyn(\VECinputSign, \timevar; \paramsROMdyn) - \TARGETstateROM_i(\timevar) \|^2
        +
        \weightREGrhs \lossREG(\paramsROMdyn)
    \end{split}
\end{equation*}
In this work, we consider two different architectures for $\ROMdyn$, namely an ODE-Net \cite{chen2018neural} and an LSTM \cite{hochreiter1997long}.
In both cases, we employ normalization layers as described in Sec.~\ref{sec:methods:normalization}.

Once both the auto-encoder $\ROMdec \circ \ROMenc$ and the dynamics network $\ROMdyn$ have been trained, the predictions on the test set are obtained as follows, for $i \in \testSamples$ and $\timevar \in \trainTIMESoutputField{i}$:
\begin{equation*}
    \APPDISCRoutputField_i(\timevar) = 
    \ROMdec(\ROMdyn(\VECinputSign, \timevar; \paramsROMdyn^*);\paramsROMdec^*),
\end{equation*}
and
\begin{equation*}
    \APPoutputField_i(\cdot,\timevar) = \discrOperatorDC(
    \ROMdec(\ROMdyn(\VECinputSign, \timevar; \paramsROMdyn^*);\paramsROMdec^*)
    ).
\end{equation*}
Here, we denote this method as AE/ODE and AE/LSTM, depending on the architecture employed for $\ROMdyn$.

We remark that the encoder $\ROMenc$ is only instrumental to recover labeled data $\TARGETstateROM_i(\OBStimevar)$ needed to train $\ROMdyn$ and it is not used to predict the output during the testing phase.
In other words, only $\ROMdyn$ and $\ROMdec$ are retained in the testing phase.
The latter observation suggests a third training stage in which, starting from the pre-trained values $(\paramsROMdyn^{*}, \paramsROMdec^{*})$, we further train the $\ROMdyn$ and $\ROMdec$ in a simultaneous manner, that is in an end-to-end (e2e) way:
\begin{equation*}
    \begin{split}
        (\paramsROMdyn^{**}, \paramsROMdec^{**}) = 
        \underset{\paramsROMdyn, \paramsROMdec}{\operatorname{argmin}} 
        &\avsum_{i \in \trainSamples}
        \avsum_{\timevar \in \trainTIMESoutputField{i}}
        \|\ROMdec(\ROMdyn(\VECinputSign, \timevar; \paramsROMdyn);\paramsROMdec) - \DISCRoutputField_i(\timevar) \|_{\DISCRoutputField}^2
        \\&+
        \weightREGrhs \lossREG(\paramsROMdyn)
        +
        \weightREGdec \lossREG(\paramsROMdec)
    \end{split}
\end{equation*}
To denote the models obtained after this third training stage, we employ the abbreviation AE/ODE-e2e and AE/LSTM-e2e.
	\end{appendices}

	\printbibliography

\end{document}